\documentclass[10pt,twocolumn,letterpaper]{article}

\usepackage[pagenumbers]{cvpr} 

\definecolor{cvprblue}{rgb}{0.21,0.49,0.74}
\usepackage[pagebackref,breaklinks,colorlinks,allcolors=cvprblue]{hyperref}
\usepackage{xcolor}
\usepackage{multirow}
\usepackage{multicol}
\usepackage{kotex}
\usepackage{color, colortbl}
\usepackage{soul}
\usepackage{inconsolata}    
\usepackage{float}      
\usepackage{placeins}   
\usepackage{hyperref}   
\usepackage[most]{tcolorbox}
\definecolor{Gray}{gray}{0.9}
\sethlcolor{Gray}                                                  
\usepackage[inline]{enumitem}
\usepackage{textcomp} 
\usepackage{tabularx,array}
\newcolumntype{C}{>{\centering\arraybackslash}X}
\def\paperID{37313} 
\def\confName{CVPR}
\def\confYear{2026}

\title{Keep it SymPL: Symbolic Projective Layout for \\Allocentric Spatial Reasoning in Vision-Language Models}

\author{
Jaeyun Jang, Seunghui Shin, Taeho Park, Hyoseok Hwang\thanks{Corresponding author} \\
Kyung Hee University, Yongin, Republic of Korea \\
{\tt\small \{yoon2926, jumin1116, katehoya, hyoseok\}@khu.ac.kr}
}

\begin{document}
\maketitle
\begin{abstract}
Perspective-aware spatial reasoning involves understanding spatial relationships from specific viewpoints—either egocentric (observer-centered) or allocentric (object-centered).
While vision–language models (VLMs) perform well in egocentric settings, their performance deteriorates when reasoning from allocentric viewpoints, where spatial relations must be inferred from the perspective of objects within the scene.
In this study, we address this underexplored challenge by introducing \textbf{Sym}bolic \textbf{P}rojective \textbf{L}ayout (SymPL), a framework that reformulates allocentric reasoning into symbolic-layout forms that VLMs inherently handle well.
By leveraging four key factors—projection, abstraction, bipartition, and localization—SymPL converts allocentric questions into structured symbolic-layout representations.
Extensive experiments demonstrate that this reformulation substantially improves performance in both allocentric and egocentric tasks, enhances robustness under visual illusions and multi-view scenarios, and that each component contributes critically to these gains.
These results show that SymPL provides an effective and principled approach for addressing complex perspective-aware spatial reasoning.

\end{abstract}    

\section{Introduction}
\label{sec:intro}

Vision–language models (VLMs) have rapidly evolved, bridging visual perception and language understanding through large-scale multimodal learning~\cite{vlm_gemini2.5, vlm_qwen2.5-vl, vlm_gpt5}. 
Early models excelled in perception-focused tasks such as visual question answering and image captioning~\cite{cite_image_captioning, cite_visual_grounding, cite_object_recognition}.
These advances have transformed VLMs from perception-oriented systems into general reasoning engines capable of understanding spatial structures and interacting with the physical world.
In particular, spatial reasoning, which involves interpreting and inferring object relationships in three-dimensional (3D) space, has become essential for embodied artificial intelligence (AI) systems, including manipulation~\cite{sp_manip, sp_manip2} and navigation~\cite{sp_navig, sp_navig2}.

Despite recent progress, the spatial reasoning capability of VLMs remains limited.
Prior studies~\cite{sr_spatialvlm, sr_spatialrgpt, sr_spatialbot, sr_sdvlm} have achieved partial success in egocentric reasoning, where relationships are interpreted from the observer’s viewpoint, but performance drops sharply in allocentric settings, which require reasoning from the perspective of objects within the scene.
This gap largely stems from the strong egocentric bias in existing training datasets~\cite{par_comfort, par_spare}, causing models to underperform when confronted with viewpoint transformations.
Such limitations hinder the use of VLMs in real-world tasks requiring multi-view or object-centered reasoning, such as autonomous driving~\cite{par_how2comm} and robotic manipulation~\cite{par_sat, par_guiding}.

\begin{figure}[t!]
    \centering
    \includegraphics[width=\linewidth]{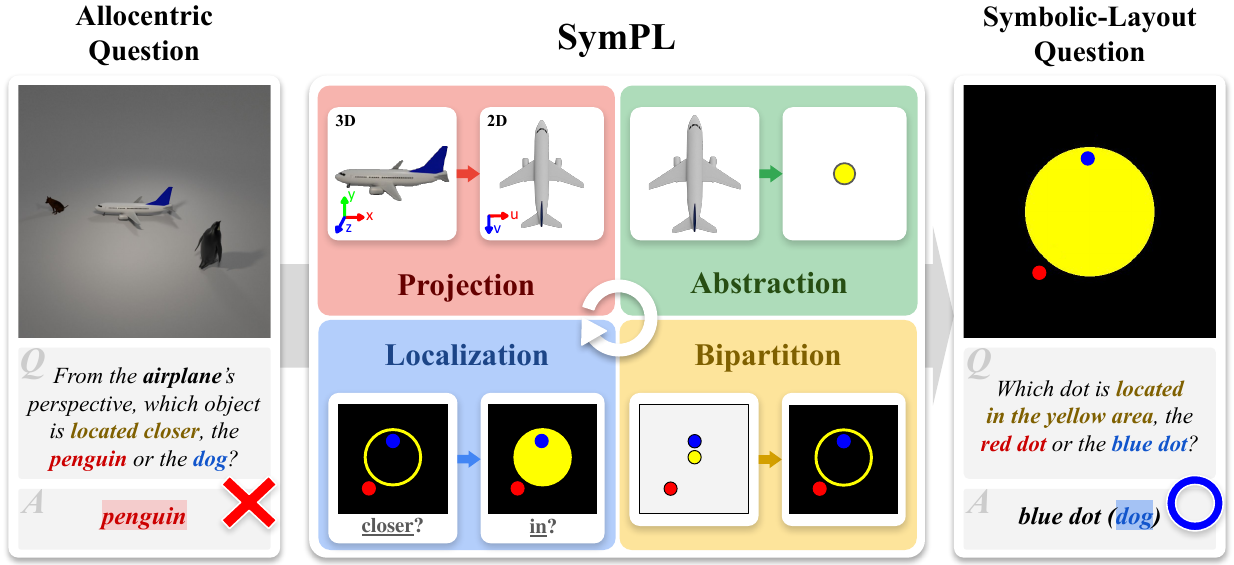} 
    \caption{SymPL reformulates allocentric questions into symbolic-layout questions using four factors-\textit{projection}, \textit{abstraction}, \textit{bipartition}, and \textit{localization}-enabling significantly improved spatial reasoning under allocentric settings.
    }
    \label{fig:introduction}
    \vspace{-5mm}
\end{figure}

Several strategies have been explored to address the limitations of allocentric spatial reasoning.
A natural approach is to train models from scratch on allocentric datasets~\cite{par_3d, par_pretrain}, mirroring how egocentric spatial skills are typically acquired.
Although such models can reach performance levels comparable to egocentric reasoning, the scarcity of allocentric data and the high computational cost make this approach impractical at scale.
Fine-tuning pre-trained VLMs offers a more feasible alternative~\cite{par_sat}, but these models often generalize poorly beyond training contexts and suffer from catastrophic forgetting when new relations are introduced.
Given these limitations, recent work has focused on improving allocentric reasoning without additional training, leveraging the inherent capabilities of large pre-trained VLMs.
General reasoning aids such as chain-of-thought (CoT)~\cite{par_cot} and visual prompting (VP)~\cite{vip_som, vip_scaffold} provide limited gains because they do not directly tackle viewpoint transformation.
More recent approach~\cite{par_apc} attempts to mitigate this challenge by transforming allocentric queries into egocentric ones—where VLMs tend to perform better—using auxiliary cues from foundation models~\cite{fm_groundingdino, fm_sam, fm_depthpro, fm_orientanything}.
However, this conversion still underuses the intrinsic reasoning capacity of VLMs, leaving substantial room for improvement.

Building on these observations, we propose a new perspective: extending beyond viewpoint conversion, we transform the problem to leverage the factors that  strongly influence VLM performance.
To this end, we first analyze which factors correlate with higher answer accuracy in VLMs, and utilize these observations to reformulate allocentric reasoning in a way that maximizes their effectiveness.
Through this analysis, we distill four fundamental factors that characterize effective spatial reasoning in VLMs:

\begin{itemize}
    
\item  \textit{Projection} : Spatial relations become easier for VLMs to process when projected onto a two-dimensional (2D) plane from an orthogonal viewpoint~\cite{vlm_view_topviewrs}.

\item  \textit{Abstraction} : Simplifying complex visual scenes into minimal, abstract symbols reduces irrelevant distractions, thereby enhancing reasoning robustness~\cite{vlm_abstract_ivispar, par_CLEVR}.

\item  \textit{Bipartition} : Spatial relationships are conveyed more intuitively when the reasoning space is minimally partitioned~\cite{vlm_bipartition_grid, vlm_bipartition_mask4align}.

\item  \textit{Localization} : Asking whether an object lies within color-coded regions, rather than reasoning about direction or distance, improves inference accuracy~\cite{vlm_localization_spatialmm, vlm_localization_colorbench}.

\end{itemize}

Building on these factors, we propose \textit{\textbf{Sym}bolic \textbf{P}rojective \textbf{L}ayout} (SymPL), which reformulates allocentric questions into symbolic layouts that align with the reasoning patterns VLMs inherently handle well.
As illustrated in Figure~\ref{fig:introduction}, SymPL estimates 3D information from the image and question, applies an orthogonal projection for relational comparison (\textit{projection}), abstracts objects into minimal symbols (\textit{abstraction}), partitions the space into two colored regions (\textit{bipartition}), and converts the query into a position-estimation task (\textit{localization}).
These steps yield a concise \textit{symbolic-layout question} that preserves essential spatial cues while filtering out redundant details, which can also be applied to egocentric spatial reasoning.
Our experiments show that symbolic-layout questions substantially improve perspective-aware spatial reasoning accuracy and consistency.
They preserve essential relational information while removing distracting cues, yielding clear gains in allocentric reasoning and also enhancing performance under egocentric, visual-illusion, and multi-view conditions.
The contributions of this work are summarized as follows:

\begin{itemize}
\item We introduce SymPL, a method that optimizes complex allocentric spatial reasoning problems into forms where VLMs naturally excel.
\item SymPL transforms allocentric questions into symbolic-layout questions using four key factors: \textit{projection}, \textit{abstraction}, \textit{bipartition}, and \textit{localization}.
\item Experiments demonstrate that SymPL improves accuracy and ensures consistent, robust perspective-aware spatial reasoning across both allocentric and egocentric settings.
\end{itemize}

\section{Related Works}
\label{sec:formatting}

\subsection{Egocentric Spatial Reasoning}
Egocentric spatial reasoning focuses on understanding spatial and geometric relationships from an observer-centered perspective~\cite{sr_whatsup, sr_mmvp, sr_cocospatial}.
Accurate reasoning of this kind relies on the alignment between visual relations in the image and their textual descriptions~\cite{sr_adaptvis}.
However, most VLMs do not enforce this alignment during training, yielding weak performance in such reasoning~\cite{vlm_llava, vlm_qwen2.5-vl}.
To address this issue and enhance this capability, several studies have been proposed. 
SpatialVLM builds a large egocentric spatial reasoning dataset with 3D information and fine-tunes models on it~\cite{sr_spatialvlm}.
SpatialRGPT injects object-level segmentation masks during training~\cite{sr_spatialrgpt}. 
SpatialBot uses a progressive pipeline to integrate multi-level spatial cues~\cite{sr_spatialbot}. 
SD-VLM learns from RGB and depth using metrically annotated 3D data~\cite{sr_sdvlm}. 
Yet, these methods reason only from the egocentric views, and their performance degrades when the questions are from the allocentric views.



\subsection{Allocentric Spatial Reasoning}

Allocentric spatial reasoning refers to the ability to understand spatial relationships from the viewpoints of other objects in the scene rather than from the observer’s perspective~\cite{par_CLEVR, par_vpt}.
This ability is essential for real-world applications that require understanding spatial relations based on surrounding objects, such as autonomous driving~\cite{par_how2comm} and robotic interaction~\cite{par_sat, par_guiding}.
Accordingly, several studies have been proposed to improve this reasoning capability.
Some approaches address this problem by training models with allocentric datasets, yet they rely on costly data such as multi-view images or videos and learn only limited positional relations~\cite{par_3d, par_sat}.
As an alternative, APC reformulates the task into an egocentric form without additional training~\cite{par_apc}, leveraging auxiliary information extracted from foundation models~\cite{fm_groundingdino, fm_sam, fm_depthpro, fm_orientanything}.
However, this conversion still falls short of leveraging the intrinsic reasoning capacity of VLMs.


\begin{figure*}[t]
    \centering
    \includegraphics[width=\linewidth]{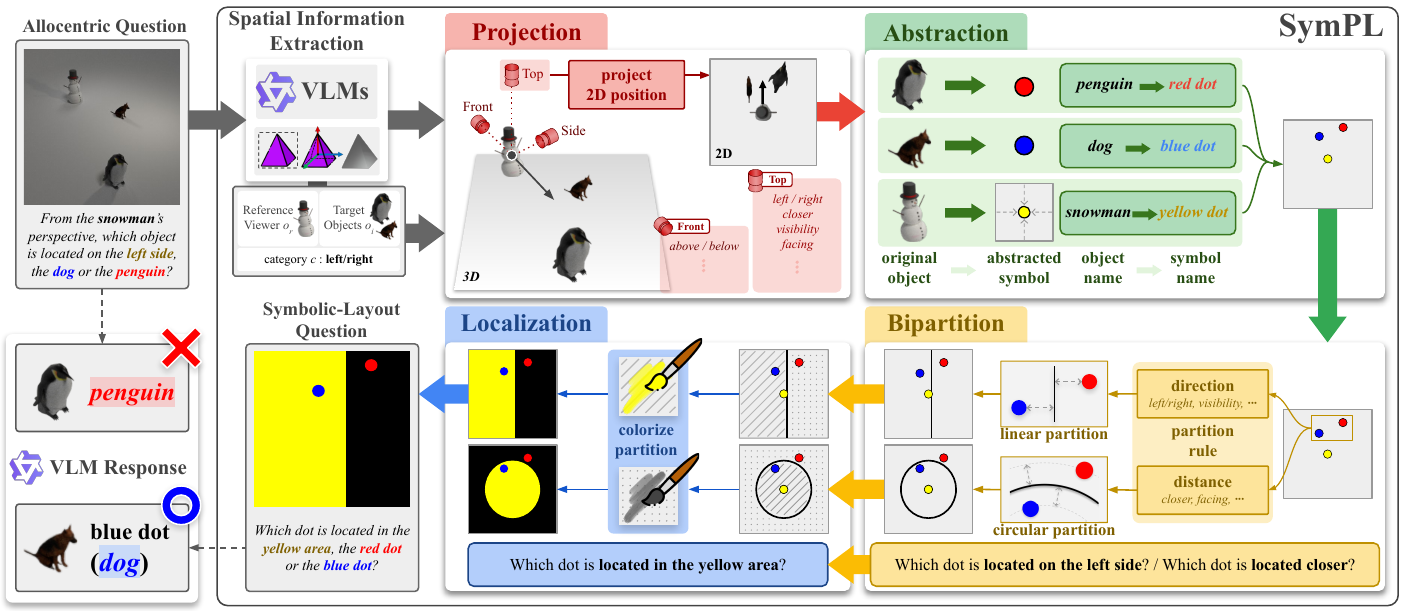}
    \caption{Overview of SymPL framework. SymPL reformulates an allocentric question into a symbolic-layout question through two stages: 1) Spatial Information Extraction and 2) Question Reformulation using four key factors — \textit{projection}, \textit{abstraction}, \textit{bipartition}, and \textit{localization}.}
    \vspace{-4mm}
    \label{fig:main_architecture}
\end{figure*}

\subsection{General Reasoning Aids for Exploiting VLMs}
Several reasoning aids have been proposed to enhance VLMs’ reasoning capabilities.
Chain-of-Thought (CoT) prompting guide models to perform complex multi-step reasoning through step-by-step examples, achieving strong performance in arithmetic and symbolic reasoning tasks~\cite{par_cot}.
Visual prompting is an annotation-based technique that helps VLMs focus on specific visual regions~\cite{vip_redcircle, vip_api, vip_fgvp, vip_vipllava, vip_pivot}.
This approach encourages the model to focus more on the highlighted regions during reasoning.
SoM prompting~\cite{vip_som} introduces a method that applies semi-transparent segmentation masks to an image to distinguish semantic regions. 
SCAFFOLD prompting~\cite{vip_scaffold} marks regularly spaced dots on an image, intuitively visualizing object positions and clarifying spatial structure for the model.
However, it remains unclear whether these approaches effectively enhance allocentric spatial reasoning.

\section{Methodology}
Our framework, SymPL, reformulates spatial reasoning problems from the allocentric view into an intuitive symbolic-layout question.
As illustrated in Figure~\ref{fig:main_architecture}, SymPL performs reasoning through two stages: Spatial Information Extraction and Question Reformulation using four key factors.
In the first stage, SymPL extracts 3D information for each object using pretrained models and a VLM. 
Next, SymPL integrates four key factors into the 3D information, namely \textit{projection}, \textit{abstraction}, \textit{bipartition}, and \textit{localization}, and generates a symbolic-layout question.
This question is used as input to a VLM instead of the original question, to indirectly infer the answer to that question.
The detailed descriptions of each stage are provided in the following sections.


\subsection{Spatial Information Extraction}
Given an allocentric question $Q$, consisting of an input image $I$ and a prompt $T$, the framework first classifies the objects that constitute $Q$.
The objects are composed of the \textit{reference viewer} $o_r$, which serves as the basis of the perspective, and the \textit{target objects} $o_i$, which are the reasoning targets.  
We perform two step reasoning processes to complete this classification.  
In the first reasoning step, the VLM extracts all object names mentioned in the prompt $T$ as a list. 
This list includes both reference viewer and target objects.
In the second reasoning step, the VLM also identifies the reference viewer from this list and constructs the final object set $O = \{o_r, o_i \mid i = 1, 2, \ldots, n\}$.  
By default, the reference viewer is explicitly specified in the prompt $T$.
However, for egocentric questions, the `camera' is designated as the reference viewer.

In the next step, we estimate the 3D coordinates of all objects.  
First, we detect the bounding boxes $B = \{b_r, b_i \mid i = 1, 2, \ldots, n\}$ for each object using GroundingDINO~\cite{fm_groundingdino} and estimate the depth map $D$ for the image $I$ using DepthPro~\cite{fm_depthpro}.  
For each object $o_j$, where $j\in\{r,1,\dots,n\}$, the corresponding bounding box $b_j$ is applied to the depth map $D$, the pixels in this region are unprojected to 3D, and the median of these points defines the object's 3D position $p_j = (x_j, y_j, z_j)$. 
During this process, we identified the area with the highest density of depth values inside the bounding box and used it to select inliers. 
We then removed outliers to reduce background, which minimized non-object regions in the mask.
To handle scale differences between the $x, y$ coordinates and the $z$ values from the depth map, we apply a correction when the scale difference exceeds a predefined threshold to minimize distortion or bias in spatial relations.

In addition, the framework estimates the facing direction vector $v_r$ of the reference viewer $o_r$ in 3D space.
First, the image $I$ is cropped according to the reference viewer’s bounding box $b_r$.
The cropped image is then passed into the OrientAnything~\cite{fm_orientanything}, which returns the facing direction vector $v_r$ for the reference viewer.
This result is combined with the object positions to form the 3D information set $U = \{v_r, p_r, p_i \mid i = 1, 2, \ldots, n\}$.

\begin{figure}[t!]
  \centering
    \includegraphics[width=\linewidth]{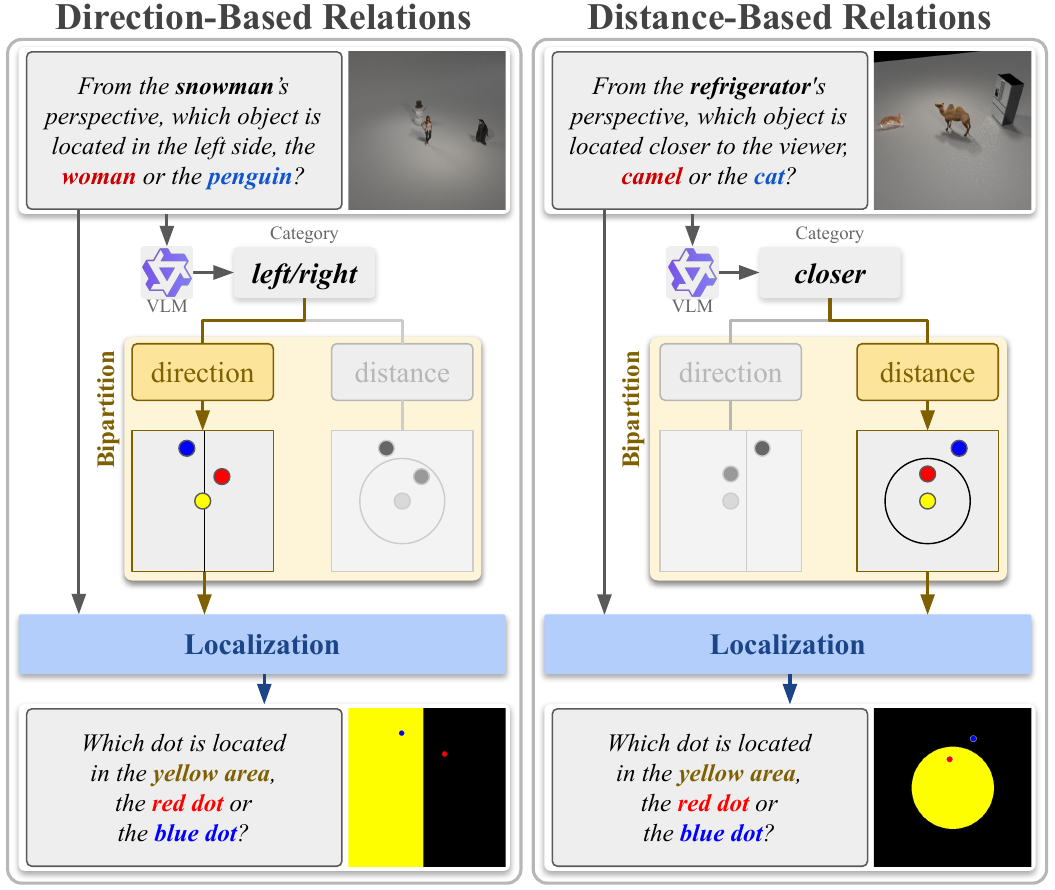} 

    \caption{Partition rule based on spatial reasoning category. Directional comparisons adopt a linear partition, while distance comparisons employ a circular one.}
    \vspace{-4mm}
    \label{fig:bipartition_rule}
\end{figure}

\subsection{Question Reformulation using four key factors}
This subsection describes how a symbolic-layout question is generated from the 3D information set $U$ through four steps: \textit{projection}, \textit{abstraction}, \textit{bipartition}, and \textit{localization}.
Each step is designed to apply the four key factors that characterize efficient spatial reasoning to allocentric questions.
Before the steps, we first feed the prompt $Q$ into the VLM to predict the spatial reasoning category $c$ that needs to be inferred.
The predicted category then serves as the guideline in each step for how the image should be reconstructed.

\noindent\textbf{Projection Step.}\enskip
The projection step first chooses an external viewpoint centered on the reference viewer $o_r$ and orthogonal to the plane where spatial relations are expressed.
We choose the top view when reasoning category $c$ is about relations on the plane on which the reference viewer stands, such as \textit{left/right}, \textit{closer}, \textit{visibility}, and \textit{facing}. 
We choose the front view for height related reasoning category such as \textit{above/below}.
Then, we project each object's 3D coordinate $p_j$ to a 2D coordinate $d_j$ in the chosen viewpoint.
When projecting the 2D coordinate, we fix the reference viewer's facing direction $v_r$ to the upward direction in the 2D plane so that the allocentric view maps to intuitive 2D spatial relations consistently.
And we also fix the reference viewer’s position $p_r$ at the center of the 2D image, so that the target object is projected onto the image relative to that coordinate.

\noindent\textbf{Abstraction Step.}\enskip
In this step, each original object is placed as an abstracted symbol based on the projected 2D coordinates $d_j$.
We use this abstraction step to represent them in a way that stabilizes VLM recognition, because new-viewpoint reconstruction can fail to retain object shapes.
During abstraction, objects are distinguished by unique colors so that the VLM recognizes symbols by color. 
We unify object shapes into featureless circles so that the VLM identifies symbols using color alone. 
The symbol for each object is then projected onto a 2D plane using $d_j$.
Additionally, each object name is reconstructed as a color–shape combination that denotes an abstract symbol.

\begin{table*}[!t]
\centering
\footnotesize
\resizebox{\linewidth}{!}{%
\fontsize{6.5}{7.5}\selectfont
\begin{tabular}{c|c|cccc|ccc}
\toprule
\multirow{2}{*}{Group} & \multirow{2}{*}{Method} & \multicolumn{4}{c|}{COMFORT\#} & \multicolumn{3}{c}{3DSRBench} \\ 
\cline{3-9}
& & left/right & closer & visibility & facing & left/right & visibility & facing \\ 
\midrule

\multirow{1}{*}{\shortstack{Random}}
& Random                 & 48.75 & 48.67 & 47.27 & 52.33 & 50.72 & 50.00 & 47.69 \\ 
\midrule
\multirow{7}{*}{General Purpose} 
& LLaVA-NeXT~\cite{vlm_llavanext}                   & 47.33 & 51.58 & 51.80 & 52.42 & 37.39 & 51.89 & 58.53 \\ 
& LLaVA-OneVision~\cite{vlm_llavaonevision}         & 47.92 & 57.83 & 51.88 & 50.42 & 36.82 & 47.53 & 61.42 \\ 
& Molmo~\cite{vlm_molmo}                            & 46.58 & 46.17 & 53.36 & \underline{53.58} & 39.97 & 50.29 & 59.39 \\ 
& Qwen2.5-VL~\cite{vlm_qwen2.5-vl}                  & 48.17 & 72.33 & 51.17 & 51.33 & 36.25 & 48.40 & 65.03 \\ 
& Cambrian-1~\cite{vlm_cambrian}                    & 41.17 & 80.42 & 50.70 & 41.33 & 40.83 & 53.63 & 67.05 \\ 
& GPT-5~\cite{vlm_gpt5}                             & 49.83 & \underline{84.25} & \underline{54.22} & 49.83 & 37.82 & 63.37 & 64.45 \\ 
& Gemini-2.5-Flash~\cite{vlm_gemini2.5}             & 38.33 & 77.83 & 52.34 & 51.58 & 38.40 & 64.10 & \textbf{72.25} \\ 
\midrule
\multirow{3}{*}{\shortstack{General\\ Reasoning Aids}} 
& Qwen2.5-VL + CoT~\cite{par_cot}                   & 43.25 & 70.75 & 50.39 & 44.67 & 33.52 & 51.74 & 63.44 \\ 
& Qwen2.5-VL + SoM~\cite{vip_som}                   & 46.58 & 67.25 & 51.88 & 46.42 & 37.54 & 45.64 & 65.61 \\ 
& Qwen2.5-VL + SCAFFOLD~\cite{vip_scaffold}         & \underline{52.17} & 71.17 & 50.39 & 47.42 & 34.81 & 51.16 & 62.72 \\ 
\midrule
\multirow{4}{*}{\shortstack{Egocentric\\Spatial Reasoning}} 
& SpatialVLM~\cite{sr_spatialvlm}                   & 46.83 & 63.67 & 50.78 & 49.58 & 39.54 & 52.76 & 59.39 \\     
& SpatialRGPT~\cite{sr_spatialrgpt}                 & 43.08 & 70.25 & 53.75 & 47.75 & 36.53 & 49.56 & 62.57 \\     
& SpatialBot~\cite{sr_spatialbot}                   & 46.33 & 58.83 & 50.08 & 53.08 & 39.54 & 47.09 & 47.69 \\     
& SD-VLM~\cite{sr_sdvlm}                            & 45.83 & 45.17 & 48.91 & 52.25 & 49.71 & 46.95 & 48.84 \\ 
\midrule
\multirow{3}{*}{\shortstack{Allocentric\\ Spatial Reasoning}} 
& SAT~\cite{par_sat}                                & 35.00 & 48.75 & 34.92 & 39.50 & 44.56 & 34.45    & 25.43 \\ 
& APC-Num~\cite{par_apc}                            & 47.83 & 52.50 & 34.14 & 36.92 & \underline{77.94} & 56.10 & 58.24 \\ 
& APC-Vis~\cite{par_apc}                            & 43.75 & 54.08 & 49.77 & 30.92 & 61.75 & \underline{71.37} & 64.60 \\ 
\midrule
\rowcolor{cvprblue!20}\multirow{1}{*}{Ours}
&  SymPL & \textbf{69.00} & \textbf{97.33} & \textbf{91.41} & \textbf{91.50} & \textbf{79.94} & \textbf{75.00} & \underline{70.95} \\ 
\bottomrule
\end{tabular}%
}
\caption{Quantitative results on allocentric questions. \textbf{Bold} indicates the best, while \underline{underline} represents the second best results.}
\label{tab:main_res}
\vspace{-4mm}
\end{table*}

\noindent\textbf{Bipartition Step.}\enskip
Next, the target object symbols are separated into a two-region layout within the abstracted image.
As illustrated in Figure~\ref{fig:bipartition_rule}, the partition boundary is either linear or circular, and the partition form is determined by the type of spatial reasoning category to be inferred.
If category $c$ encodes directional comparisons, we use a linear partition so that different directions fall into distinct regions. 
For the \textit{left/right} category, assuming the reference viewer’s forward direction is projected upward in the image plane, we employ a vertical partition to separate left from right. 
For the \textit{visibility} category, which determines whether the target object lies in front of or behind the reference viewer, we use a horizontal partition to separate front and back.
Conversely, if category $c$ requires comparing distances with respect to a specific location, we introduce a circular partition centered at that location to make distance differences visually distinguishable. 
For the \textit{closer} category, the goal is to select the object closer to the reference viewer, so we partition the space using a circular boundary centered at the object of interest. 
For the \textit{facing} category, which favors objects near the viewer’s facing axis, we construct a circular partition centered at a point on that axis. 
Overall, this partition rule captures the key geometric cues and enable effective visualization of diverse spatial relations in the image space.

\noindent\textbf{Localization Step.}\enskip
In the last step, questions about relative spatial relations are reformulated as a localization problem.
First, we fill the bipartitioned regions with different colors.
The colors used here are distinct from the target object symbol colors defined in the abstraction step.
The painted colors serve to visually encode linguistic expressions of position. 
For example, if the reasoning category $c$ is \textit{left/right} and the abstracted image assigns yellow to the left region and black to the right region, then the positional expression ‘left’ can be represented by the visual cue ‘yellow.’ 
Accordingly, once the space is partitioned in this way, the positional relations mentioned in the input prompt $T$ can likewise be converted into expressions about colors. 
Moreover, in this case, relative spatial questions like ‘located on the left side’ can be reduced to a localization problem such as ‘located in’ the color region that corresponds to that position.
The resulting image–prompt pair becomes a symbolic-layout question $Q^*$ that VLMs can effectively reason.

\section{Experiments}
In this section, we evaluated our framework, SymPL, across various spatial reasoning tasks and analyzed the utility of the symbolic layout question.
To this end, we constructed five benchmark datasets and conducted experiments using diverse VLMs as baselines.  
In all experiments, we used the Qwen2.5-VL~\cite{vlm_qwen2.5-vl} for all reasoning in our framework.

\begin{figure*}[!t]
    \centering
    \includegraphics[width=\linewidth]{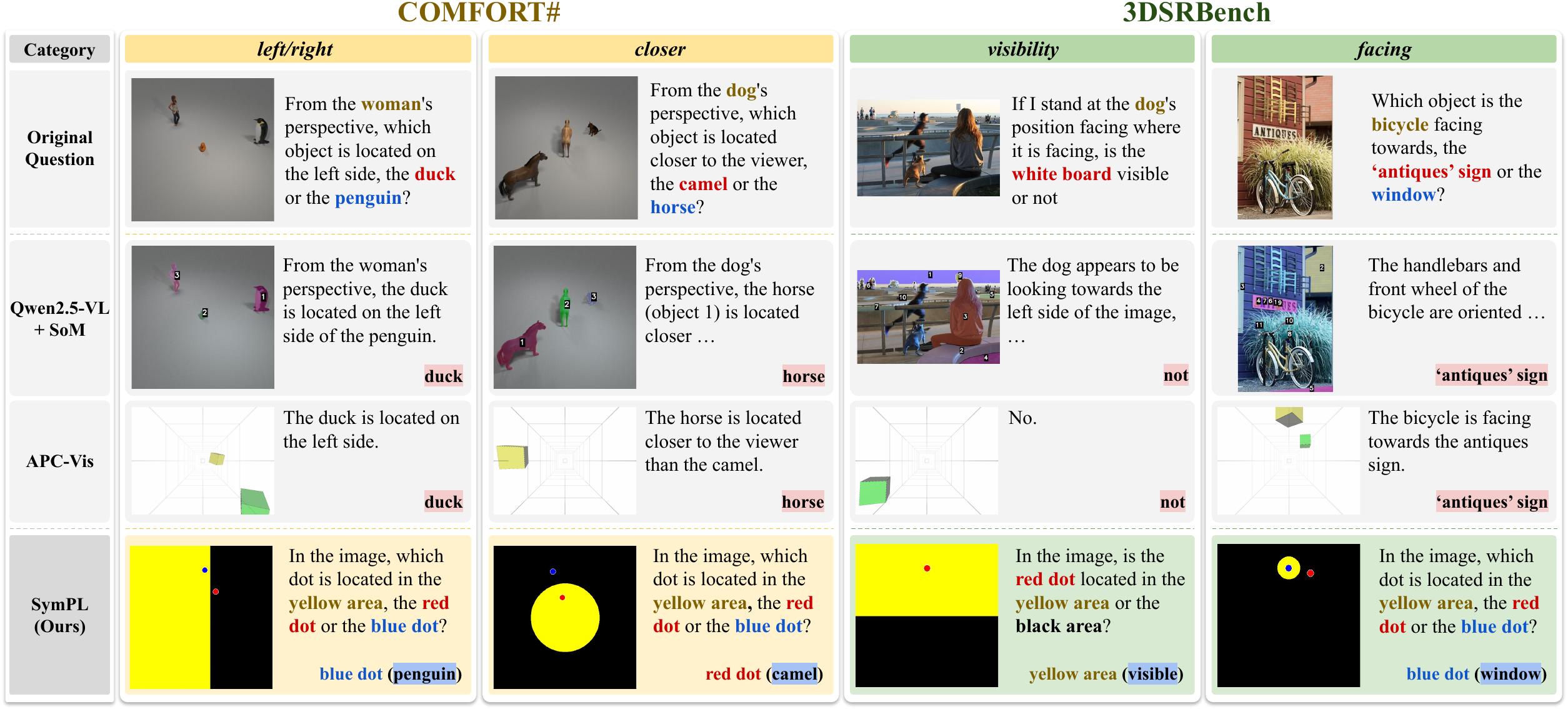}
    \caption{Allocentric spatial reasoning examples. Qwen2.5-VL + SoM and APC-Vis exhibited limited allocentric spatial reasoning performance across various categories. In contrast, our SymPL effectively handled allocentric questions by reformulating them into symbolic-layout questions.}
    \label{fig:qualitative_result}
    \vspace{-4mm}
\end{figure*}

\subsection{Experiment Setup}
\subsubsection{Baseline VLMs}
The reasoning performance of the SymPL framework was compared with that of various VLMs, and the models were categorized into four groups according to their roles and characteristics.
The first group, referred to as General Purpose, includes both open-source and API-based VLMs~\cite{vlm_llavanext, vlm_llavaonevision, vlm_qwen2.5-vl, vlm_molmo, vlm_cambrian, vlm_gpt5, vlm_gemini2.5}. 
These models represent general purpose VLMs with standard reasoning performances.  
Next, the General Reasoning Aids group includes a CoT method and two VP methods~\cite{par_cot, vip_som, vip_scaffold}.
We incorporated these methods into the Qwen2.5-VL model.
The Egocentric Spatial Reasoning group consists of four baselines, each fine-tuned for egocentric questions~\cite{sr_spatialvlm, sr_spatialrgpt, sr_spatialbot, sr_sdvlm}. 
The last group, referred to as Allocentric Spatial Reasoning, consists of three kinds of methods designed to solve allocentric questions~\cite{par_sat, par_apc}.
Additionally, we included a Random baseline that returns answers randomly to serve as a reference for comparison.

\subsubsection{Datasets}
To evaluate the performance of our model across diverse spatial reasoning tasks, we conducted experiments using five types of datasets. 
The detailed process for dataset construction is provided in the Supplementary Material.

\noindent \textbf{COMFORT\#.} Focusing on allocentric spatial reasoning, this dataset is generated using a Blender-based synthetic dataset creation pipeline~\cite{par_comfort}.
Each image is generated using an object set randomly selected from twelve types of object assets.
The dataset consists of four categories: \textit{left/right}, \textit{closer}, \textit{visibility}, and \textit{facing}.

\noindent \textbf{3DSRBench.} As a real-world benchmark for allocentric spatial reasoning, 3DSRBench is used in our experiments.
We extracted only the categories that require considering allocentric view, namely \textit{left/right}, \textit{visibility}, and \textit{facing}, for evaluation. 
The \textit{visibility} category was adapted from the original dataset’s \textit{front/behind} category.

\noindent \textbf{COCOSPATIAL.} Serving as a real-world benchmark, COCOSPATIAL targets general egocentric spatial reasoning tasks. 
We constructed our benchmark using two categories from this dataset: \textit{left/right} and \textit{above/below}.

\noindent \textbf{COMFORT VI.} Built with the COMFORT\# pipeline, COMFORT VI comprises spatial reasoning tasks under visual illusions.
A visual illusion is a size-induced misperception of spatial relations.
We constructed scenarios using colored balls of varying sizes and generated tasks per scenario, covered \textit{left/right} for allocentric question and \textit{front/behind} and \textit{closer} for egocentric question.

\noindent \textbf{COMFORT Multi.} This dataset is designed to assess the model’s consistency across viewpoints and was constructed using the COMFORT\# pipeline.
In this dataset, we captured 20 multi-view images per scene by moving the camera on a viewer-centered sphere with 72° azimuth and 15° polar steps, for 10 scenes per category.
The dataset consists of four categories: \textit{left/right}, \textit{closer}, \textit{visibility}, and \textit{facing}.

\subsection{Evaluation on Allocentric Questions}


\textbf{Evaluation on COMFORT\#.}\enskip
As shown in Table~\ref{tab:main_res}, we evaluated whether the SymPL framework performs accurate reasoning across a range of allocentric questions.
On COMFORT\#, SymPL achieved \textit{left/right} 69.00\%, \textit{closer} 97.33\%, \textit{visibility} 91.41\%, and \textit{facing} 91.50\%, and outperformed all prior methods across categories. 
These results showed that the SymPL framework effectively performed allocentric spatial reasoning across multiple categories based on symbolic-layout questions.
However, most baseline VLMs performed at or below the random baseline, and even GPT-5, which achieved the highest performance in the \textit{closer} category as 84.25\%, also performed worse in other categories.
These findings indicated the limitations of VLMs’ allocentric spatial reasoning and highlighted the importance of the proposed framework.


\begin{table}[t!]
\centering
\resizebox{0.8\columnwidth}{!}{%
\begin{tabular}{c|ccc}
\toprule
\multirow{2}{*}{Method} & \multicolumn{2}{c}{COCOSPATIAL}                     \\ \cline{2-3} 
                        & left/right     & above/below    \\ \midrule
Random                  & 50.25          & 50.00          \\ \midrule
LLaVa-NeXT              & \underline{88.67}          & 76.00          \\
LLaVa-OneVision         & 87.67          & 82.67          \\
Molmo                   & 81.17          & 72.17          \\
Qwen2.5-VL              & 86.33          & 78.75          \\
Cambrian-1              & 67.00          & 67.50          \\
GPT-5                    & 80.17          & 84.25          \\
Gemini-2.5-Flash        & 88.58          & \underline{92.42}          \\ \midrule
Qwen2.5-VL + CoT        & 71.33          & 69.50          \\
Qwen2.5-VL + SoM        & 79.58          & 74.17          \\
Qwen2.5-VL + SCAFFOLD   & 84.33          & 81.92          \\ \midrule
SpatialVLM              & 78.92          & 71.00          \\
SpatialRGPT             & 84.00          & 85.25          \\
SpatialBot              & 84.50          & 83.50          \\
SD-VLM                  & 71.25          & 51.33          \\ \midrule
SAT                     & 39.00          & 38.58          \\
APC-Num                 & 49.00          & 27.00          \\
APC-Vis                 & 49.92          & 54.17          \\ \midrule
\rowcolor{cvprblue!20} SymPL (Ours)   & \textbf{89.83}          & \textbf{94.33}   \\ \bottomrule
\end{tabular}
}
\caption{Quantitative results on egocentric questions. \textbf{Bold} indicates the best, while \underline{underline} represents the second best results.}
\label{tab:sr_res}
\vspace{-5mm}
\end{table}

\noindent\textbf{Evaluation on 3DSRBench.}\enskip
On 3DSRBench, SymPL achieved the highest accuracy in \textit{left/right} at 79.94\% and in \textit{visibility} at 75.00\%, and placed second in \textit{facing} with 70.95\%.
Notably, many baselines tended to perform over 10\% worse than the random baseline in the \textit{left/right} category, showing that VLMs were biased toward an egocentric view.
In the case of the APC series, the robustness of reasoning performance was low across various categories.
Similarly, Gemini-2.5-Flash, which achieved the highest performance of 72.25\% in the \textit{facing} category, showed only 38.40\% in the \textit{left/right} category, underscoring weak cross-category consistency.
In contrast, SymPL’s consistent gains across all categories demonstrated robustness to diverse spatial relations and validated the effectiveness of the symbolic-layout questions. 
Qualitative examples are provided in Figure~\ref{fig:qualitative_result}.

\begin{table}[!t]
\centering
\resizebox{\columnwidth}{!}{%
\begin{tabular}{c|ccc}
\toprule
\multirow{3}{*}{Method} & \multicolumn{3}{c}{COMFORT VI}                           \\ \cline{2-4} 
                        & \multicolumn{1}{c|}{Allocentric}        & \multicolumn{2}{c}{Egocentric} \\ \cline{2-4} 
                        & \multicolumn{1}{c|}{left/right} & front/behind  & closer \\ \midrule
Random                  & \multicolumn{1}{c|}{49.69}      & 51.75         & 49.13  \\ \midrule
LLaVa-NeXT              & \multicolumn{1}{c|}{58.19}      & 58.75         & 89.13  \\
LLaVa-OneVision         & \multicolumn{1}{c|}{72.00}      & 95.88         & \underline{98.75}  \\
Molmo                   & \multicolumn{1}{c|}{44.56}      & 93.88         & 74.50  \\
Qwen2.5-VL              & \multicolumn{1}{c|}{57.00}      & 82.63         & 47.25  \\
Cambrian-1              & \multicolumn{1}{c|}{35.06}      & 79.25         & 87.25  \\
GPT-5                   & \multicolumn{1}{c|}{43.63}      & \underline{99.75}         & 89.13  \\
Gemini-2.5-Flash        & \multicolumn{1}{c|}{54.63}      & 55.50         & 44.00  \\ \midrule
Qwen2.5-VL + CoT        & \multicolumn{1}{c|}{40.94}      & 97.25         & 23.50  \\
Qwen2.5-VL + SoM        & \multicolumn{1}{c|}{46.56}      & 44.25         & 46.13  \\
Qwen2.5-VL + SCAFFOLD   & \multicolumn{1}{c|}{58.06}      & 85.00         & 68.50  \\ \midrule
SpatialVLM              & \multicolumn{1}{c|}{44.00}      & 92.13         & 80.75  \\
SpatialRGPT             & \multicolumn{1}{c|}{42.31}      & 64.88         & 61.38  \\
SpatialBot              & \multicolumn{1}{c|}{52.25}      & 78.25         & 87.00  \\
SD-VLM                  & \multicolumn{1}{c|}{50.00}      & 49.13         & 46.75  \\ \midrule
SAT                     & \multicolumn{1}{c|}{29.44}      & 49.63         & 47.63  \\
APC-Num                 & \multicolumn{1}{c|}{\underline{84.31}}      & 27.88         & 23.00  \\
APC-Vis                 & \multicolumn{1}{c|}{76.75}      & 36.25         & 42.25  \\ \midrule
\rowcolor{cvprblue!20} SymPL (Ours)            & \multicolumn{1}{c|}{\textbf{95.38}}      & \textbf{100.00}           & \textbf{100.00}    \\ \bottomrule
\end{tabular}
}
\caption{Quantitative results on perspective-aware reasoning under visual illusions. \textbf{Bold} indicates the best, while \underline{underline} represents the second best results.}
\vspace{-5mm}
\label{tab:vi_res}
\end{table}

\subsection{Applying SymPL to Egocentric Questions}
Next, the applicability of the SymPL framework for egocentric spatial reasoning was evaluated using COCOSPATIAL.
The results are presented in Table~\ref{tab:sr_res}. In this experiment, SymPL achieved 89.83\% in \textit{left/right} and 94.33\% in \textit{above/below}, which exceeded the best baselines. 
The results demonstrated that the symbolic-layout question yielded marked performance improvements even for egocentric questions.
With most baselines having surpassed 70\%, this showed that overall VLM performance was biased toward egocentric questions rather than allocentric ones.
By contrast, APC-Num and APC-Vis showed a clear drop in accuracy, because the models are biased toward allocentric questions, misclassifying the camera viewpoint as allocentric rather than egocentric and generating incorrect scenes.
These trends showed the effectiveness of the SymPL, which supports both allocentric and egocentric reasoning.

\subsection{Assessing Perspective-Aware Reasoning under Visual Illusions}

To assess spatial reasoning in more diverse scenarios, we conducted additional experiments using a dataset featuring induced visual illusions.
In this setup, \textit{left/right} is evaluated as an allocentric question, while \textit{front/behind} and \textit{closer} are egocentric. 
As shown in Table~\ref{tab:vi_res}, SymPL achieved 100\% on \textit{front/behind} and \textit{closer} and 95.38\% on \textit{left/right}, which demonstrated robust reasoning even under visual illusions. 
Additionally, among baseline VLMs, the APC series specialized in allocentric spatial reasoning, while the others were stronger in the egocentric setting.
This suggested that most VLMs were biased toward one perspective or the other.
By contrast, our framework again showed robust reasoning under visual illusions.

\subsection{Evaluation of Viewpoint-Aware Consistency}
An additional experiment was conducted to assess whether our approach yields consistent reasoning across images of the same scene captured from different viewpoints. 
As baselines, methods that exploit existing VLMs more effectively were considered, including CoT, SoM, SCAFFOLD, APC-Num, and APC-Vis.
As shown in Table~\ref{tab:multi_res}, SymPL achieved the highest success rate across all categories.
These results indicated that our method supported robust allocentric reasoning that was invariant to the image-capture viewpoint.

\begin{table}[!t]
\centering
\resizebox{\columnwidth}{!}{%
\begin{tabular}{c|cccc}
\toprule
Method                & left/right & closer & visibility & facing \\ \midrule
Qwen2.5-VL            & 67.50   & \underline{70.50}       & 58.50       & \underline{57.50}   \\ \midrule
Qwen2.5-VL + CoT      & 57.50   & 66.00       & 60.50       & 50.50   \\
Qwen2.5-VL + SoM      & \underline{72.50}   & 51.50       & \underline{61.50}       & 54.50   \\
Qwen2.5-VL + SCAFFOLD & 66.50   & 53.50       & 54.00       & 55.00   \\ \midrule
APC-Num               & 44.50   & 28.50       & 55.00       & 23.50       \\ 
APC-Vis               & 53.00   & 31.50       & \underline{61.50}       & 16.50       \\  \midrule
\rowcolor{cvprblue!20} SymPL (Ours)          & \textbf{76.00}   & \textbf{96.50}       & \textbf{86.00}       & \textbf{74.00}  \\
\bottomrule
\end{tabular}
}
\caption{Quantitative results on viewpoint-aware consistency across multiple views. \textbf{Bold} indicates the best, while \underline{underline} represents the second best results.}
\vspace{-3mm}
\label{tab:multi_res}
\end{table}

\subsection{Ablation Studies}

\subsubsection{Analysis of Each Key Factor}
\begin{figure}[t!]
  \centering

  \begin{subfigure}{0.48\linewidth}
    \centering
    \includegraphics[width=\linewidth]{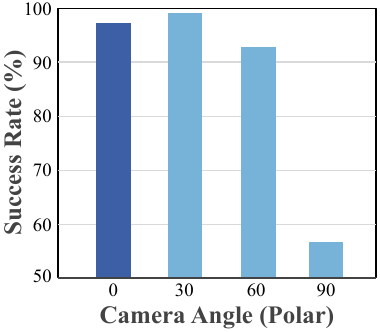}
    \captionsetup{margin={1.6em,0pt}}
    \caption{}
    \label{fig:ablation_four_factors:a}
  \end{subfigure}\hfill
  \begin{subfigure}{0.48\linewidth}
    \centering
    \includegraphics[width=\linewidth]{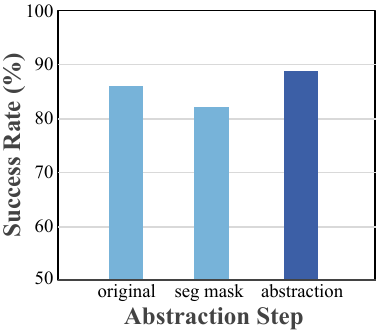}
    \captionsetup{margin={1.6em,0pt}}
    \caption{}
    \label{fig:ablation_four_factors:b}
  \end{subfigure}

  \vspace{0.4em}

  \begin{subfigure}{0.48\linewidth}
    \centering
    \includegraphics[width=\linewidth]{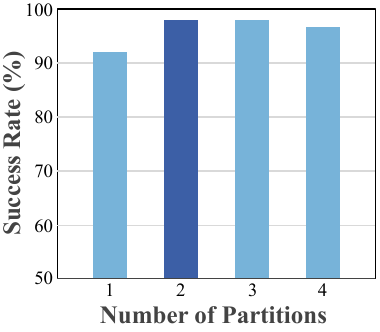}
    \captionsetup{margin={1.6em,0pt}}
    \caption{}
    \label{fig:ablation_four_factors:c}
  \end{subfigure}\hfill
  \begin{subfigure}{0.48\linewidth}
    \centering
    \includegraphics[width=\linewidth]{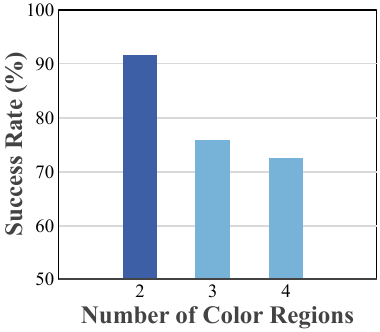}
    \captionsetup{margin={1.6em,0pt}}
    \caption{}
    \label{fig:ablation_four_factors:d}
  \end{subfigure}

  \caption{Ablation results of each key factor. (a) \textit{projection}, (b) \textit{abstraction}, (c) \textit{bipartition}, (d) \textit{localization}. The darker bar indicates the configuration used in SymPL.}
    \vspace{-5mm}
  \label{fig:ablation_four_factors}
\end{figure}

We analyzed the contribution of each of the four key factors to verify whether they have an effective impact on VLM reasoning. 
The experimental results are shown in Figure~\ref{fig:ablation_four_factors}.
\textbf{Projection.}\enskip
Viewpoint-dependent tendencies in spatial relation understanding were investigated by moving the camera from a front view to a top view while keeping the scene fixed.
We built the scene using COMFORT\# pipeline.
We then analyzed the trend of reasoning performance changes according to viewpoint in the \textit{above/below} category. 
As shown in Figure~\ref{fig:ablation_four_factors:a}, reasoning performance on \textit{above/below} decreased as the camera approached a top view, indicating that selecting a viewpoint that matched the spatial relationships was crucial.
\\
\textbf{Abstraction.}\enskip
The effect of abstraction was examined in the \textit{closer} category by comparing performance when the original images were annotated with segmentation masks against performance when they were converted into abstract images.
Figure~\ref{fig:ablation_four_factors:b} showed that expressing positional relationships in an abstracted form has a positive effect on reasoning compared to segmentation mask on the original image.
\\
\textbf{Bipartition.}\enskip
The impact of bipartition was evaluated in the \textit{closer} category by measuring spatial reasoning performance as a function of the number of partitions.
Figure~\ref{fig:ablation_four_factors:c} illustrated that the presence of partitions had a positive effect on reasoning, but the number of partitions made little difference in performance. 
\\
\textbf{Localization.}\enskip
We analyzed reasoning performance as the number of regions distinguished by different colors increased using the \textit{closer} category.
In Figure~\ref{fig:ablation_four_factors:d}, performance dropped greatly as the number of regions to consider increased,
confirming that partitioning a region into two parts and assigning different colors to them yields the notable gains.

\subsubsection{Ablation on the Effectiveness of the Key Factor}
An ablation study was conducted to examine how the performance on symbolic-layout questions changes as four key factors are added step by step. 
The experimental setup and results were summarized in Table~\ref{tab:ab_res}. 
Starting from Setting 1, we sequentially incorporate \textit{projection}, \textit{abstraction}, \textit{bipartition}, and \textit{localization}, yielding the full symbolic-layout question at Setting 5. 
For evaluation, we randomly sampled 100 instances per category and tested five general purpose VLMs: Qwen2.5-VL, GPT-5, LLaVA-NeXT, LLaVA-OneVision, and Molmo. 
From the experimental results, Setting 1 exhibited the lowest performance in most categories, and performance tended to improve as the stages progressed. Under Setting 5, all experiments uniformly achieved 100\%.
These results indicated that the four key factors acted synergistically and combining them improved reasoning performance.

\subsection{Error Breakdown}
\begin{figure}[t!]
  \centering
    \includegraphics[width=\linewidth]{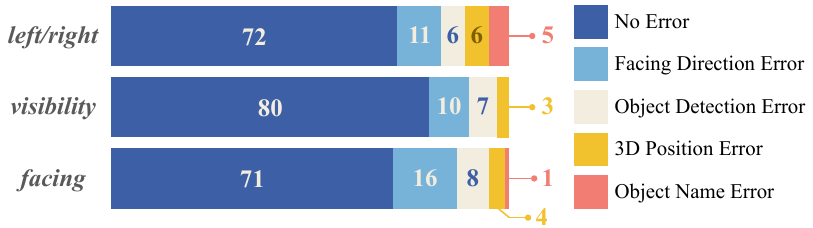} 

    \caption{Error breakdown in SymPL’s reasoning pipeline.}
    \vspace{-4mm}
    \label{fig:error_breakdown}
\end{figure}

We analyzed a manual error breakdown to assess failure cases in SymPL’s reasoning pipeline, using inference results from 100 randomly sampled instances per 3DSRBench category.
Figure~\ref{fig:error_breakdown} showed that the most frequent error across categories involved misestimating the reference viewer’s facing-direction vector. 
Other common errors included incorrect object detection, inaccurate 3D coordinate estimation, and misidentification of object names specified in the prompt. 
Most failures appeared to arise from reliance on foundation models, and reasoning failures on the symbolic-layout question were not observed in this analysis.

\begin{table}[!t]
\centering
\resizebox{\columnwidth}{!}{%
\begin{tabular}{c|c|c|c|c|cccc}
\toprule
Setting & P & A & B & L & \textit{left/right} & \textit{closer} & \textit{visibility} & \textit{facing} \\ \midrule
1       &            &             &             &             & 46.60  & 63.80 & 52.00 & 52.80 \\ \midrule
2       & \checkmark &             &             &             & 89.20  & 64.80 & 51.20 & 52.00 \\ \midrule
3       & \checkmark & \checkmark  &             &             & 96.40  & 81.00 & 90.80 & 100.00 \\ \midrule
4       & \checkmark & \checkmark  & \checkmark  &             & 97.00  & 91.00 & 84.60 & 100.00 \\ \midrule
\rowcolor{cvprblue!20} 5 (Ours)       & \checkmark & \checkmark  & \checkmark  & \checkmark  & 100.00 & 100.00 & 100.00 & 100.00 \\ \bottomrule
\end{tabular}
}
\caption{Ablation results on the effectiveness of four key factors: \textbf{P}rojection, \textbf{A}bstraction, \textbf{B}ipartition, and \textbf{L}ocalization. Results show the average success rate of five general-purpose VLMs for each category: \textit{left/right}, \textit{closer}, \textit{visibility}, and \textit{facing}.}
\label{tab:ab_res}
\end{table}

\section{Conclusion}

In this paper, we introduce SymPL, a framework that optimizes complex spatial reasoning problem into forms where VLMs excel effectively.
SymPL reformulates questions into intuitive symbolic-layout questions based on four key factors, enabling effective reasoning.
Our experiments show that symbolic-layout questions significantly improve both allocentric and egocentric spatial reasoning.
These results demonstrate that SymPL provides an effective and principled approach for addressing complex perspective-aware spatial reasoning.



{
    \small
    \bibliographystyle{ieeenat_fullname}
    \bibliography{main}

\begin{thebibliography}{51}
\providecommand{\natexlab}[1]{#1}
\providecommand{\url}[1]{\texttt{#1}}
\expandafter\ifx\csname urlstyle\endcsname\relax
  \providecommand{\doi}[1]{doi: #1}\else
  \providecommand{\doi}{doi: \begingroup \urlstyle{rm}\Url}\fi

\bibitem[Bai et~al.(2025)Bai, Chen, Liu, Wang, Ge, Song, Dang, Wang, Wang, Tang, Zhong, Zhu, Yang, Li, Wan, Wang, Ding, Fu, Xu, Ye, Zhang, Xie, Cheng, Zhang, Yang, Xu, and Lin]{vlm_qwen2.5-vl}
Shuai Bai, Keqin Chen, Xuejing Liu, Jialin Wang, Wenbin Ge, Sibo Song, Kai Dang, Peng Wang, Shijie Wang, Jun Tang, Humen Zhong, Yuanzhi Zhu, Mingkun Yang, Zhaohai Li, Jianqiang Wan, Pengfei Wang, Wei Ding, Zheren Fu, Yiheng Xu, Jiabo Ye, Xi Zhang, Tianbao Xie, Zesen Cheng, Hang Zhang, Zhibo Yang, Haiyang Xu, and Junyang Lin.
\newblock Qwen2.5-vl technical report.
\newblock \emph{arXiv preprint arXiv:2502.13923}, 2025.

\bibitem[Beckham et~al.(2023)Beckham, Weiss, Golemo, Honari, Nowrouzezahrai, and Pal]{par_CLEVR}
Christopher Beckham, Martin Weiss, Florian Golemo, Sina Honari, Derek Nowrouzezahrai, and Christopher Pal.
\newblock Visual question answering from another perspective: Clevr mental rotation tests.
\newblock \emph{Pattern Recognition}, 136:\penalty0 109209, 2023.

\bibitem[Bochkovskii et~al.(2025)Bochkovskii, Delaunoy, Germain, Santos, Zhou, Richter, and Koltun]{fm_depthpro}
Aleksei Bochkovskii, Ama\"{e}l Delaunoy, Hugo Germain, Marcel Santos, Yichao Zhou, Stephan~R. Richter, and Vladlen Koltun.
\newblock Depth pro: Sharp monocular metric depth in less than a second.
\newblock In \emph{International Conference on Learning Representations}, 2025.

\bibitem[Cai et~al.(2024{\natexlab{a}})Cai, Liu, Mustikovela, Meyer, Chai, Park, and Lee]{vip_vipllava}
Mu Cai, Haotian Liu, Siva~Karthik Mustikovela, Gregory~P. Meyer, Yuning Chai, Dennis Park, and Yong~Jae Lee.
\newblock Making large multimodal models understand arbitrary visual prompts.
\newblock In \emph{IEEE Conference on Computer Vision and Pattern Recognition}, 2024{\natexlab{a}}.

\bibitem[Cai et~al.(2024{\natexlab{b}})Cai, Ponomarenko, Yuan, Li, Yang, Dong, and Zhao]{sr_spatialbot}
Wenxiao Cai, Yaroslav Ponomarenko, Jianhao Yuan, Xiaoqi Li, Wankou Yang, Hao Dong, and Bo Zhao.
\newblock Spatialbot: Precise spatial understanding with vision language models.
\newblock \emph{IEEE International Conference on Robotics and Automation (ICRA)}, 2024{\natexlab{b}}.

\bibitem[Chen et~al.(2024)Chen, Xu, Kirmani, Ichter, Sadigh, Guibas, and Xia]{sr_spatialvlm}
Boyuan Chen, Zhuo Xu, Sean Kirmani, Brian Ichter, Dorsa Sadigh, Leonidas Guibas, and Fei Xia.
\newblock Spatialvlm: Endowing vision-language models with spatial reasoning capabilities.
\newblock In \emph{Proceedings of the IEEE/CVF Conference on Computer Vision and Pattern Recognition (CVPR)}, pages 14455--14465, 2024.

\bibitem[Chen et~al.(2025{\natexlab{a}})Chen, Lou, Cao, Guo, Fan, Wu, Yang, Ma, and Ye]{sr_sdvlm}
Pingyi Chen, Yujing Lou, Shen Cao, Jinhui Guo, Lubin Fan, Yue Wu, Lin Yang, Lizhuang Ma, and Jieping Ye.
\newblock Sd-vlm: Spatial measuring and understanding with depth-encoded vision-language models.
\newblock \emph{Advances in Neural Information Processing Systems}, 2025{\natexlab{a}}.

\bibitem[Chen et~al.(2025{\natexlab{b}})Chen, Zhu, Zhou, Zhang, Gao, Niebles, Geva, He, Wu, and Li]{sr_adaptvis}
Shiqi Chen, Tongyao Zhu, Ruochen Zhou, Jinghan Zhang, Siyang Gao, Juan~Carlos Niebles, Mor Geva, Junxian He, Jiajun Wu, and Manling Li.
\newblock Why is spatial reasoning hard for vlms? an attention mechanism perspective on focus areas.
\newblock In \emph{International conference on machine learning}, 2025{\natexlab{b}}.

\bibitem[Cheng et~al.(2024)Cheng, Yin, Fu, Guo, Yang, Kautz, Wang, and Liu]{sr_spatialrgpt}
An-Chieh Cheng, Hongxu Yin, Yang Fu, Qiushan Guo, Ruihan Yang, Jan Kautz, Xiaolong Wang, and Sifei Liu.
\newblock Spatialrgpt: Grounded spatial reasoning in vision-language models.
\newblock In \emph{NeurIPS}, 2024.

\bibitem[Comanici et~al.(2025)Comanici, Bieber, Schaekermann, Pasupat, Sachdeva, Dhillon, Blistein, Ram, Zhang, Rosen, et~al.]{vlm_gemini2.5}
Gheorghe Comanici, Eric Bieber, Mike Schaekermann, Ice Pasupat, Noveen Sachdeva, Inderjit Dhillon, Marcel Blistein, Ori Ram, Dan Zhang, Evan Rosen, et~al.
\newblock Gemini 2.5: Pushing the frontier with advanced reasoning, multimodality, long context, and next generation agentic capabilities.
\newblock \emph{arXiv preprint arXiv:2507.06261}, 2025.

\bibitem[Crespo et~al.(2020)Crespo, Castillo, Mozos, and Barber]{sp_navig}
Jonathan Crespo, Jose~Carlos Castillo, Oscar~Martinez Mozos, and Ramon Barber.
\newblock Semantic information for robot navigation: A survey.
\newblock \emph{Applied Sciences}, 10\penalty0 (2):\penalty0 497, 2020.

\bibitem[Deitke et~al.(2025)Deitke, Clark, Lee, Tripathi, Yang, Park, Salehi, Muennighoff, Lo, Soldaini, et~al.]{vlm_molmo}
Matt Deitke, Christopher Clark, Sangho Lee, Rohun Tripathi, Yue Yang, Jae~Sung Park, Mohammadreza Salehi, Niklas Muennighoff, Kyle Lo, Luca Soldaini, et~al.
\newblock Molmo and pixmo: Open weights and open data for state-of-the-art vision-language models.
\newblock In \emph{Proceedings of the Computer Vision and Pattern Recognition Conference}, pages 91--104, 2025.

\bibitem[Del~Sette et~al.(2022)Del~Sette, Bindemann, and Ferguson]{par_vpt}
Paola Del~Sette, Markus Bindemann, and Heather~J Ferguson.
\newblock Visual perspective-taking in complex natural scenes.
\newblock \emph{Quarterly Journal of Experimental Psychology}, 75\penalty0 (8):\penalty0 1541--1551, 2022.

\bibitem[Gao et~al.(2024)Gao, Sarkar, Xia, Xiao, Wu, Ichter, Majumdar, and Sadigh]{sp_manip2}
Jensen Gao, Bidipta Sarkar, Fei Xia, Ted Xiao, Jiajun Wu, Brian Ichter, Anirudha Majumdar, and Dorsa Sadigh.
\newblock Physically grounded vision-language models for robotic manipulation.
\newblock In \emph{2024 IEEE International Conference on Robotics and Automation (ICRA)}, pages 12462--12469. IEEE, 2024.

\bibitem[Guo et~al.(2024)Guo, Xu, Yao, Cui, Ni, Ge, Chua, Liu, and Huang]{cite_object_recognition}
Zonghao Guo, Ruyi Xu, Yuan Yao, Junbo Cui, Zanlin Ni, Chunjiang Ge, Tat-Seng Chua, Zhiyuan Liu, and Gao Huang.
\newblock Llava-uhd: an lmm perceiving any aspect ratio and high-resolution images.
\newblock In \emph{European Conference on Computer Vision}, pages 390--406. Springer, 2024.

\bibitem[Izadi et~al.(2025)Izadi, Banayeeanzade, Askari, Rahimiakbar, Vahedi, Hasani, and Soleymani~Baghshah]{vlm_bipartition_grid}
Amirmohammad Izadi, Mohammadali Banayeeanzade, Fatemeh Askari, Ali Rahimiakbar, Mohammad~Mahdi Vahedi, Hosein Hasani, and Mahdieh Soleymani~Baghshah.
\newblock Visual structures help visual reasoning: Addressing the binding problem in {LVLM}s.
\newblock In \emph{The Thirty-Ninth Conference on Neural Information Processing Systems (NeurIPS 2025)}, 2025.
\newblock Poster.

\bibitem[Kamath et~al.(2023)Kamath, Hessel, and Chang]{sr_whatsup}
Amita Kamath, Jack Hessel, and Kai-Wei Chang.
\newblock What{'}s ``up'' with vision-language models? investigating their struggle with spatial reasoning.
\newblock In \emph{Proceedings of the 2023 Conference on Empirical Methods in Natural Language Processing}, pages 9161--9175, Singapore, 2023. Association for Computational Linguistics.

\bibitem[Kang et~al.(2025)Kang, Kim, Kim, and Hwang]{cite_visual_grounding}
Seil Kang, Jinyeong Kim, Junhyeok Kim, and Seong~Jae Hwang.
\newblock Your large vision-language model only needs a few attention heads for visual grounding.
\newblock In \emph{Proceedings of the Computer Vision and Pattern Recognition Conference}, pages 9339--9350, 2025.

\bibitem[Kirillov et~al.(2023)Kirillov, Mintun, Ravi, Mao, Rolland, Gustafson, Xiao, Whitehead, Berg, Lo, Doll{\'a}r, and Girshick]{fm_sam}
Alexander Kirillov, Eric Mintun, Nikhila Ravi, Hanzi Mao, Chloe Rolland, Laura Gustafson, Tete Xiao, Spencer Whitehead, Alexander~C. Berg, Wan-Yen Lo, Piotr Doll{\'a}r, and Ross Girshick.
\newblock Segment anything.
\newblock \emph{Proceedings of the IEEE/CVF International Conference on Computer Vision (ICCV)}, 2023.

\bibitem[Kojima et~al.(2022)Kojima, Gu, Reid, Matsuo, and Iwasawa]{par_cot}
Takeshi Kojima, Shixiang~Shane Gu, Machel Reid, Yutaka Matsuo, and Yusuke Iwasawa.
\newblock Large language models are zero-shot reasoners.
\newblock \emph{Advances in neural information processing systems}, 35:\penalty0 22199--22213, 2022.

\bibitem[Lee et~al.(2025{\natexlab{a}})Lee, Je, Park, Uy, Guibas, and Sung]{par_apc}
Phillip~Y Lee, Jihyeon Je, Chanho Park, Mikaela~Angelina Uy, Leonidas Guibas, and Minhyuk Sung.
\newblock Perspective-aware reasoning in vision-language models via mental imagery simulation.
\newblock In \emph{Proceedings of the IEEE/CVF International Conference on Computer Vision (ICCV)}, 2025{\natexlab{a}}.

\bibitem[Lee et~al.(2025{\natexlab{b}})Lee, Choi, Kang, Kim, Park, and Shim]{par_3d}
Seonho Lee, Jiho Choi, Inha Kang, Jiwook Kim, Junsung Park, and Hyunjung Shim.
\newblock 3d-aware vision-language models fine-tuning with geometric distillation.
\newblock In \emph{Findings of the Association for Computational Linguistics: EMNLP 2025}, 2025{\natexlab{b}}.

\bibitem[Lei et~al.(2024)Lei, Yang, Chen, Li, and Liu]{vip_scaffold}
Xuanyu Lei, Zonghan Yang, Xinrui Chen, Peng Li, and Yang Liu.
\newblock Scaffolding coordinates to promote vision-language coordination in large multi-modal models.
\newblock \emph{Conference on Computational Linguistics}, 2024.

\bibitem[Li et~al.(2024{\natexlab{a}})Li, Zhang, Guo, Zhang, Li, Zhang, Zhang, Li, Liu, and Li]{vlm_llavaonevision}
Bo Li, Yuanhan Zhang, Dong Guo, Renrui Zhang, Feng Li, Hao Zhang, Kaichen Zhang, Yanwei Li, Ziwei Liu, and Chunyuan Li.
\newblock Llava-onevision: Easy visual task transfer.
\newblock \emph{arXiv preprint arXiv:2408.03326}, 2024{\natexlab{a}}.

\bibitem[Li et~al.(2024{\natexlab{b}})Li, Zhang, Zhou, Collier, Korhonen, and Vuli{\'c}]{vlm_view_topviewrs}
Chengzu Li, Caiqi Zhang, Han Zhou, Nigel Collier, Anna Korhonen, and Ivan Vuli{\'c}.
\newblock Topviewrs: Vision-language models as top-view spatial reasoners.
\newblock In \emph{Proceedings of the 2024 Conference on Empirical Methods in Natural Language Processing}, pages 1786--1807, Miami, Florida, USA, 2024{\natexlab{b}}. Association for Computational Linguistics.

\bibitem[Li et~al.(2024{\natexlab{c}})Li, Zhang, Geng, Geng, Long, Shen, Zhang, Liu, and Dong]{sp_manip}
Xiaoqi Li, Mingxu Zhang, Yiran Geng, Haoran Geng, Yuxing Long, Yan Shen, Renrui Zhang, Jiaming Liu, and Hao Dong.
\newblock Manipllm: Embodied multimodal large language model for object-centric robotic manipulation.
\newblock In \emph{Proceedings of the IEEE/CVF Conference on Computer Vision and Pattern Recognition}, pages 18061--18070, 2024{\natexlab{c}}.

\bibitem[Liang et~al.(2025)Liang, Li, Fan, Li, Nguyen, Cobbina, Bhardwaj, Chen, Liu, and Zhou]{vlm_localization_colorbench}
Yijun Liang, Ming Li, Chenrui Fan, Ziyue Li, Dang Nguyen, Kwesi~Adu Cobbina, Shweta Bhardwaj, Jiuhai Chen, Fuxiao Liu, and Tianyi Zhou.
\newblock Colorbench: Can vlms see and understand the colorful world? a comprehensive benchmark for color perception, reasoning, and robustness.
\newblock In \emph{Advances in Neural Information Processing Systems (NeurIPS)}, 2025.

\bibitem[Liu et~al.(2023{\natexlab{a}})Liu, Li, Wu, and Lee]{vlm_llava}
Haotian Liu, Chunyuan Li, Qingyang Wu, and Yong~Jae Lee.
\newblock Visual instruction tuning.
\newblock \emph{Advances in neural information processing systems}, 36:\penalty0 34892--34916, 2023{\natexlab{a}}.

\bibitem[Liu et~al.(2024)Liu, Li, Li, Li, Zhang, Shen, and Lee]{vlm_llavanext}
Haotian Liu, Chunyuan Li, Yuheng Li, Bo Li, Yuanhan Zhang, Sheng Shen, and Yong~Jae Lee.
\newblock Llava-next: Improved reasoning, ocr, and world knowledge, 2024.

\bibitem[Liu et~al.(2023{\natexlab{b}})Liu, Zeng, Ren, Li, Zhang, Yang, Li, Yang, Su, Zhu, et~al.]{fm_groundingdino}
Shilong Liu, Zhaoyang Zeng, Tianhe Ren, Feng Li, Hao Zhang, Jie Yang, Chunyuan Li, Jianwei Yang, Hang Su, Jun Zhu, et~al.
\newblock Grounding dino: Marrying dino with grounded pre-training for open-set object detection.
\newblock \emph{European Conference on Computer Vision}, 2023{\natexlab{b}}.

\bibitem[Mayer et~al.(2025)Mayer, Ballout, Jassim, Nezami, and Bruni]{vlm_abstract_ivispar}
Julius Mayer, Mohamad Ballout, Serwan Jassim, Farbod~Nosrat Nezami, and Elia Bruni.
\newblock i{VISPAR} {---} an interactive visual-spatial reasoning benchmark for {VLM}s.
\newblock In \emph{Proceedings of the 2025 Conference on Empirical Methods in Natural Language Processing}, pages 26745--26769, Suzhou, China, 2025. Association for Computational Linguistics.

\bibitem[Nasiriany et~al.(2024)Nasiriany, Xia, Yu, Xiao, Liang, Dasgupta, Xie, Driess, Wahid, Xu, Vuong, Zhang, Lee, Lee, Xu, Kirmani, Zhu, Zeng, Hausman, Heess, Finn, Levine, and Ichter]{vip_pivot}
Soroush Nasiriany, Fei Xia, Wenhao Yu, Ted Xiao, Jacky Liang, Ishita Dasgupta, Annie Xie, Danny Driess, Ayzaan Wahid, Zhuo Xu, Quan Vuong, Tingnan Zhang, Tsang-Wei~Edward Lee, Kuang-Huei Lee, Peng Xu, Sean Kirmani, Yuke Zhu, Andy Zeng, Karol Hausman, Nicolas Heess, Chelsea Finn, Sergey Levine, and Brian Ichter.
\newblock Pivot: Iterative visual prompting elicits actionable knowledge for vlms.
\newblock \emph{International conference on machine learning}, 2024.

\bibitem[Ogezi and Shi(2025)]{par_spare}
Michael Ogezi and Freda Shi.
\newblock {S}pa{RE}: Enhancing spatial reasoning in vision-language models with synthetic data.
\newblock In \emph{Proceedings of the 63rd Annual Meeting of the Association for Computational Linguistics (Volume 1: Long Papers)}, pages 7855--7875, Vienna, Austria, 2025. Association for Computational Linguistics.

\bibitem[{OpenAI}(2025)]{vlm_gpt5}
{OpenAI}.
\newblock Gpt-5 system card.
\newblock Technical report, OpenAI, 2025.
\newblock Version as of Aug 2025.

\bibitem[Ranasinghe et~al.(2024)Ranasinghe, Shukla, Poursaeed, Ryoo, and Lin]{sr_cocospatial}
Kanchana Ranasinghe, Satya~Narayan Shukla, Omid Poursaeed, Michael~S Ryoo, and Tsung-Yu Lin.
\newblock Learning to localize objects improves spatial reasoning in visual-llms.
\newblock In \emph{Proceedings of the IEEE/CVF Conference on Computer Vision and Pattern Recognition}, pages 12977--12987, 2024.

\bibitem[Ray et~al.(2025)Ray, Duan, Brown, Tan, Bashkirova, Hendrix, Ehsani, Kembhavi, Plummer, Krishna, Zeng, and Saenko]{par_sat}
Arijit Ray, Jiafei Duan, Ellis Brown, Reuben Tan, Dina Bashkirova, Rose Hendrix, Kiana Ehsani, Aniruddha Kembhavi, Bryan~A. Plummer, Ranjay Krishna, Kuo-Hao Zeng, and Kate Saenko.
\newblock Sat: Dynamic spatial aptitude training for multimodal language models.
\newblock In \emph{Proceedings of the Conference on Language Modeling (COLM 2025)}, 2025.

\bibitem[Shiri et~al.(2024)Shiri, Guo, Far, Yu, Haffari, and Li]{vlm_localization_spatialmm}
Fatemeh Shiri, Xiao-Yu Guo, Mona~Golestan Far, Xin Yu, Gholamreza Haffari, and Yuan-Fang Li.
\newblock An empirical analysis on spatial reasoning capabilities of large multimodal models.
\newblock In \emph{Proceedings of the 2024 Conference on Empirical Methods in Natural Language Processing}, pages 21440--21455, Miami, Florida, USA, 2024. Association for Computational Linguistics.

\bibitem[Shtedritski et~al.(2023)Shtedritski, Rupprecht, and Vedaldi]{vip_redcircle}
Aleksandar Shtedritski, Christian Rupprecht, and Andrea Vedaldi.
\newblock What does clip know about a red circle? visual prompt engineering for vlms.
\newblock In \emph{Proceedings of the IEEE/CVF International Conference on Computer Vision}, pages 11987--11997, 2023.

\bibitem[Tong et~al.(2024{\natexlab{a}})Tong, Brown, Wu, Woo, IYER, Akula, Yang, Yang, Middepogu, Wang, et~al.]{vlm_cambrian}
Peter Tong, Ellis Brown, Penghao Wu, Sanghyun Woo, Adithya Jairam~Vedagiri IYER, Sai~Charitha Akula, Shusheng Yang, Jihan Yang, Manoj Middepogu, Ziteng Wang, et~al.
\newblock Cambrian-1: A fully open, vision-centric exploration of multimodal llms.
\newblock \emph{Advances in Neural Information Processing Systems}, 37:\penalty0 87310--87356, 2024{\natexlab{a}}.

\bibitem[Tong et~al.(2024{\natexlab{b}})Tong, Liu, Zhai, Ma, LeCun, and Xie]{sr_mmvp}
Shengbang Tong, Zhuang Liu, Yuexiang Zhai, Yi Ma, Yann LeCun, and Saining Xie.
\newblock Eyes wide shut? exploring the visual shortcomings of multimodal llms.
\newblock In \emph{Proceedings of the IEEE/CVF Conference on Computer Vision and Pattern Recognition}, pages 9568--9578, 2024{\natexlab{b}}.

\bibitem[Wang et~al.(2024)Wang, Zhang, Pang, Du, Zhao, and Zhao]{fm_orientanything}
Zehan Wang, Ziang Zhang, Tianyu Pang, Chao Du, Hengshuang Zhao, and Zhou Zhao.
\newblock Orient anything: Learning robust object orientation estimation from rendering 3d models.
\newblock \emph{International conference on machine learning}, 2024.

\bibitem[Werner et~al.(1997)Werner, Krieg-Br{\"u}ckner, Mallot, Schweizer, and Freksa]{sp_navig2}
Steffen Werner, Bernd Krieg-Br{\"u}ckner, Hanspeter~A Mallot, Karin Schweizer, and Christian Freksa.
\newblock Spatial cognition: The role of landmark, route, and survey knowledge in human and robot navigation1.
\newblock In \emph{Informatik’97 Informatik als Innovationsmotor: 27. Jahrestagung der Gesellschaft f{\"u}r Informatik Aachen, 24.--26. September 1997}, pages 41--50. Springer, 1997.

\bibitem[Yang et~al.(2023{\natexlab{a}})Yang, Yang, Wang, Liu, Xu, Yin, Zhai, and Zhang]{par_how2comm}
Dingkang Yang, Kun Yang, Yuzheng Wang, Jing Liu, Zhi Xu, Rongbin Yin, Peng Zhai, and Lihua Zhang.
\newblock How2comm: Communication-efficient and collaboration-pragmatic multi-agent perception.
\newblock \emph{Advances in Neural Information Processing Systems}, 36:\penalty0 25151--25164, 2023{\natexlab{a}}.

\bibitem[Yang et~al.(2023{\natexlab{b}})Yang, Zhang, Li, Zou, Li, and Gao]{vip_som}
Jianwei Yang, Hao Zhang, Feng Li, Xueyan Zou, Chunyuan Li, and Jianfeng Gao.
\newblock Set-of-mark prompting unleashes extraordinary visual grounding in gpt-4v, 2023{\natexlab{b}}.

\bibitem[Yang et~al.(2023{\natexlab{c}})Yang, Wang, Li, Wang, and Yang]{vip_fgvp}
Lingfeng Yang, Yueze Wang, Xiang Li, Xinlong Wang, and Jian Yang.
\newblock Fine-grained visual prompting.
\newblock \emph{Advances in Neural Information Processing Systems}, 36:\penalty0 24993--25006, 2023{\natexlab{c}}.

\bibitem[Yang et~al.(2025)Yang, Garrett, Fox, Lozano-P{\'e}rez, and Kaelbling]{par_guiding}
Zhutian Yang, Caelan Garrett, Dieter Fox, Tom{\'a}s Lozano-P{\'e}rez, and Leslie~Pack Kaelbling.
\newblock Guiding long-horizon task and motion planning with vision language models.
\newblock In \emph{2025 IEEE International Conference on Robotics and Automation (ICRA)}, pages 16847--16853. IEEE, 2025.

\bibitem[Ye et~al.(2025)Ye, Zeng, Li, Li, and Fan]{cite_image_captioning}
Qinghao Ye, Xianhan Zeng, Fu Li, Chunyuan Li, and Haoqi Fan.
\newblock Painting with words: Elevating detailed image captioning with benchmark and alignment learning.
\newblock In \emph{Proceedings of the International Conference on Learning Representations (ICLR)}, 2025.
\newblock Accepted at ICLR 2025.

\bibitem[Yu et~al.(2024)Yu, Yu, and Wang]{vip_api}
Runpeng Yu, Weihao Yu, and Xinchao Wang.
\newblock Api: Attention prompting on image for large vision-language models.
\newblock In \emph{European Conference on Computer Vision}, 2024.

\bibitem[Zhang et~al.(2024)Zhang, Huang, Xie, and Zhang]{vlm_bipartition_mask4align}
Haoquan Zhang, Ronggang Huang, Yi Xie, and Huaidong Zhang.
\newblock Mask4align: Aligned entity prompting with color masks for multi-entity localization problems.
\newblock In \emph{Proceedings of the IEEE/CVF Conference on Computer Vision and Pattern Recognition}, pages 13373--13383, 2024.

\bibitem[Zhang et~al.(2025)Zhang, Hu, Lee, Shi, Kordjamshidi, Chai, and Ma]{par_comfort}
Zheyuan Zhang, Fengyuan Hu, Jayjun Lee, Freda Shi, Parisa Kordjamshidi, Joyce Chai, and Ziqiao Ma.
\newblock Do vision-language models represent space and how? evaluating spatial frame of reference under ambiguities.
\newblock In \emph{The Thirteenth International Conference on Learning Representations}, 2025.

\bibitem[Zheng et~al.(2025)Zheng, Huang, Li, and Wang]{par_pretrain}
Duo Zheng, Shijia Huang, Yanyang Li, and Liwei Wang.
\newblock Learning from videos for 3d world: Enhancing mllms with 3d vision geometry priors.
\newblock In \emph{Advances in Neural Information Processing Systems}, 2025.

\end{thebibliography}
}


\end{document}


\onecolumn
\maketitle

\appendix
\renewcommand{\thesection}{\Alph{section}}

\renewcommand{\thefigure}{S\arabic{figure}}
\renewcommand{\thetable}{S\arabic{table}}
\setcounter{figure}{0}
\setcounter{table}{0}

\makeatletter
\renewcommand{\theHfigure}{S\arabic{figure}}
\renewcommand{\theHtable}{S\arabic{table}}
\makeatother


\section{SymPL Framework}
\subsection{Implementation Details}
\begin{figure}[H]
    \centering
    \includegraphics[width=0.9\linewidth]{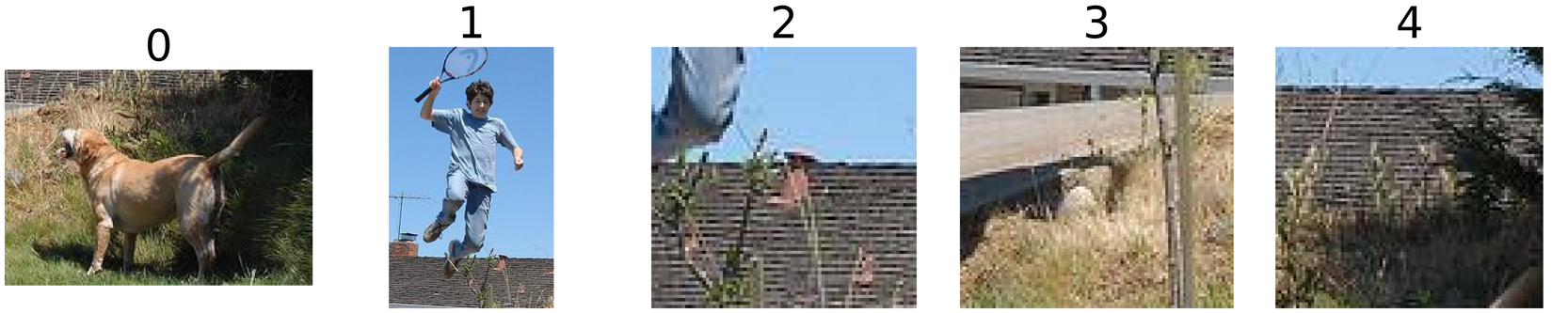}
    \caption{Candidate cropped images fed to the VLM. The target object is the \textbf{dog}. Each image is indexed at the top according to its confidence score ranking, and the VLM returns the index of the image that best matches the target object through reasoning.
    }
    \label{fig:object_detection}
\end{figure}

\paragraph{Object Detection.}
To extract 3D object information from an image, we used GroundingDINO~\cite{fm_groundingdino} to obtain 2D bounding boxes.  
When using only the bounding box with the highest confidence score, the model frequently detected incorrect objects.  
To address this issue, we selected the top $n$ bounding boxes based on their confidence scores and generated cropped images for each of them.  
As shown in Figure~\ref{fig:object_detection}, these $n$ cropped images were arranged sequentially with their corresponding ranks.  
To include minimal contextual information around the target object, we expanded the bounding box by a fixed $margin$ in all directions before cropping the image.
We then prompted the VLM to identify the image that best matched the target object, and used the bounding box corresponding to the selected cropped image as the final detection result.  
In all experiments, we set $n = 5$ and $margin = 30$ for object detection.

\paragraph{3D Position Estimation.}
To estimate the precise 3D coordinates of each object, we used the bounding box from GroundingDINO and the depth map predicted by DepthPro~\cite{fm_depthpro}.  
First, we cropped the corresponding region on the depth map using the bounding box and extracted the depth values within the masked area.  
Then, using the intrinsic parameters predicted by DepthPro, we unprojected each pixel coordinate within the mask into 3D space to generate a 3D point cloud.  
Next, we split the range of depth values into up to 20 bins of equal width and selected the bin that contained the most depth samples.
Finally, we filtered depth values to those within a fixed ratio of the selected bin center and computed their median, which we used as the final estimate of the object’s 3D coordinate.
In all experiments, we set the ratio to $0.12$ for filtering depth values.

The accuracy of the estimated 3D coordinates depended on the intrinsic parameters predicted by DepthPro. 
When these parameters were inaccurate, the $z$ value of the 3D coordinates could become much larger or smaller than the $x$ and $y$ values. 
This issue was particularly critical in egocentric spatial reasoning, where the raw $z$ values were directly used for \textit{projection}.
To reduce this distortion, we applied an additional correction to the $z$ values before determining the final 3D coordinates. 
We first computed a $z$-axis scale as the mean absolute $z$ value of the estimated 3D positions of all objects. 
We then computed an $x$–$y$ scale as the mean distance of these 3D positions from the origin in the $x$–$y$ plane. 
If the ratio between these two scales exceeded a predefined threshold, we multiplied all $z$ values by a fixed scaling factor determined by this threshold so that the $z$-axis scale did not become excessively larger or smaller than the $x$–$y$ scale.
In the experiments, we set the threshold to $10$ for $z$-axis correction.

\section{Experiments}
\subsection{Hardware Settings}
All main experiments were conducted on a shared server using a single NVIDIA RTX A6000 GPU. 
As an exception, the APC series was run on a local machine with two NVIDIA RTX 3090 GPUs, because APC-Vis image rendering was not supported on the shared server. 
For the additional ablation studies, we used both NVIDIA RTX A6000 and NVIDIA RTX 6000 Ada Generation GPUs across the study, while each subfigure and table was composed of results obtained on a single, consistent GPU model chosen from these two.

\subsection{Dataset Configurations}
\label{sec:suppl_b_1}
We evaluated spatial reasoning abilities, including allocentric spatial reasoning, using five processed datasets: COMFORT\#, 3DSRBench, COCOSPATIAL, COMFORT VI, and COMFORT Multi. 
All experiments were conducted using the evaluation toolkit provided by VLMEvalKit~\cite{eval_vlmevalkit}. 
The preprocessing procedure for each dataset is as follows.

\begin{figure}[H]
    \centering
    \includegraphics[width=0.9\linewidth]{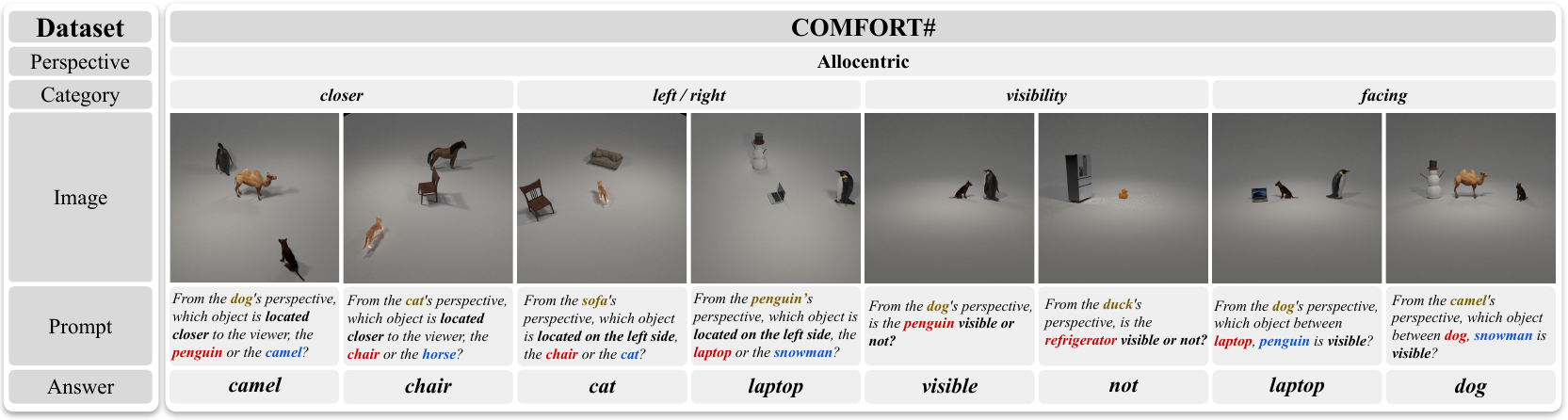}
    \caption{Examples of COMFORT\#. This dataset consists of four spatial relation categories: \textit{closer}, \textit{left/right}, \textit{visibility} and \textit{facing}.
    }
    \label{fig:dataset_sharp}
\end{figure}

\noindent \textbf{COMFORT\#.} As shown in Figure~\ref{fig:dataset_sharp}, COMFORT\# is a dataset constructed to evaluate each model's allocentric spatial reasoning ability. 
We built the dataset in a Blender-based simulation environment~\cite{par_comfort} by composing scenes from a set of 12 assets (couch, chair, dog, duck, penguin, laptop, woman, cat, refrigerator, horse, camel, and snowman), randomly sampling objects from this list for each scene.
The dataset consisted of four spatial reasoning categories: \textit{closer}, \textit{left/right}, \textit{visibility}, and \textit{facing}. For \textit{visibility}, each scene contained two objects, while the other three categories used three objects per scene.

In the \textit{closer} category, objects were arranged in a line such that the reference viewer was located at one end, and questions asked which object is closer to the reference viewer. 
In the \textit{left/right} category, we placed two objects in front of a reference viewer object and constructed questions that require inferring which object is on the left from the reference viewer's perspective. 
In the \textit{visibility} category, we randomly configured the target object to be either in front of or behind the reference viewer and constructed questions that require determining whether the target object was visible to the reference viewer in each case. 
Finally, in the \textit{facing} category, objects were again arranged linearly, but the reference viewer was positioned at the center, and questions asked which object the reference viewer was facing.
For all scenes, we injected noise into the camera position, object positions, and the facing directions of objects (except for the reference viewer), in order to generate diverse scenes.

\begin{figure}[H]
    \centering
    \includegraphics[width=0.9\linewidth]{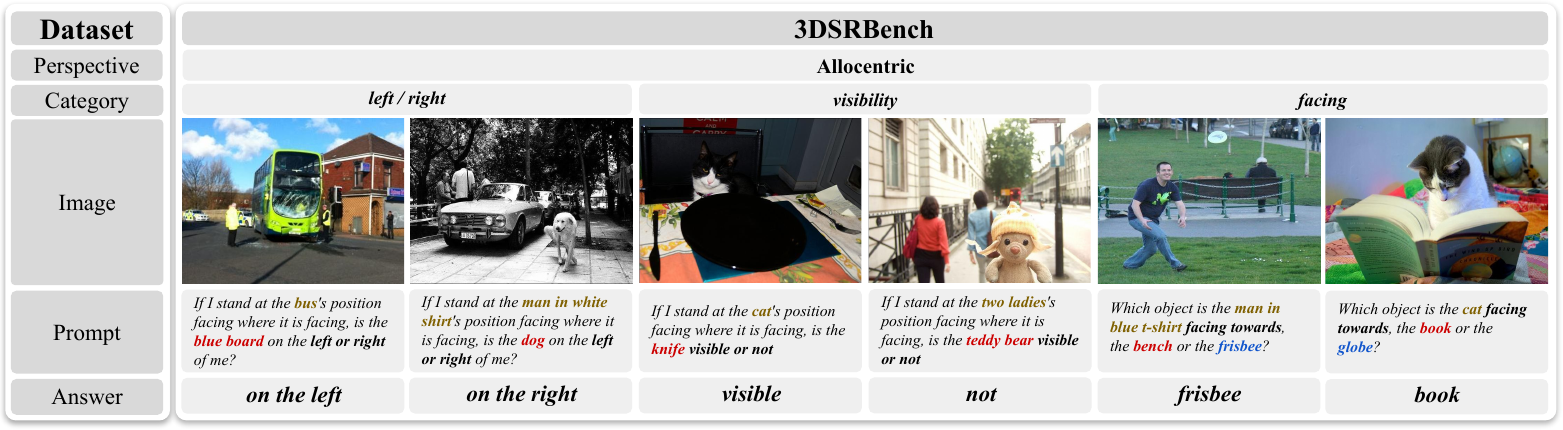}
    \caption{Examples of 3DSRBench. This dataset consists of three spatial relation categories: \textit{left/right}, \textit{visibility} and \textit{facing}.
    }
    \label{fig:dataset_3dsr}
\end{figure}

\noindent \textbf{3DSRBench.} This dataset is a real-world dataset composed of allocentric spatial reasoning questions covering various spatial relationship categories~\cite{eval_3dsr}.  
As illustrated in Figure~\ref{fig:dataset_3dsr}, we constructed an allocentric spatial reasoning dataset using the \textit{left/right}, \textit{visibility}, and \textit{facing} categories from this dataset.  
The \textit{left/right} and \textit{facing} categories were used directly from the original dataset.
For the \textit{visibility} category, we modified the existing \textit{front/behind} category such that questions with the answer \textit{front} were labeled as \textit{visible}, and those with the answer \textit{behind} were labeled as \textit{not}.

\begin{figure}[H]
    \centering
    \includegraphics[width=0.9\linewidth]{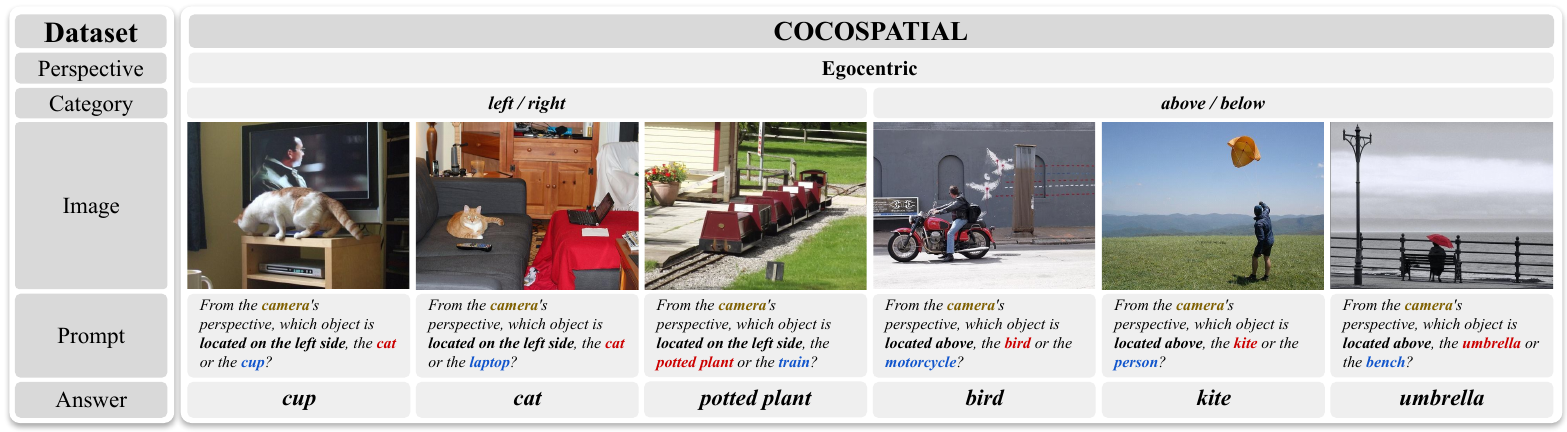}
    \caption{Examples of COCOSPATIAL. This dataset consists of two spatial relation categories: \textit{left/right} and \textit{above/below}.
    }
    \label{fig:dataset_coco}
\end{figure}

\noindent \textbf{COCOSPATIAL.} This dataset is a real-world dataset that classifies spatial relations between various object pairs based on the positions of their bounding boxes in each image~\cite{eval_cocospatial}.  
We refined this dataset to construct an egocentric spatial reasoning dataset that consisted of the \textit{left/right} and \textit{above/below} categories (see Figure~\ref{fig:dataset_coco}).  
Specifically, for each image, we randomly selected one object pair from the \textit{good pairs} with clearly defined spatial relations to form a question.  
Each question was designed to ask which object is located on the left side for the \textit{left/right} category or on the upper side for the \textit{above/below} category.

\begin{figure}[H]
    \centering
    \includegraphics[width=0.9\linewidth]{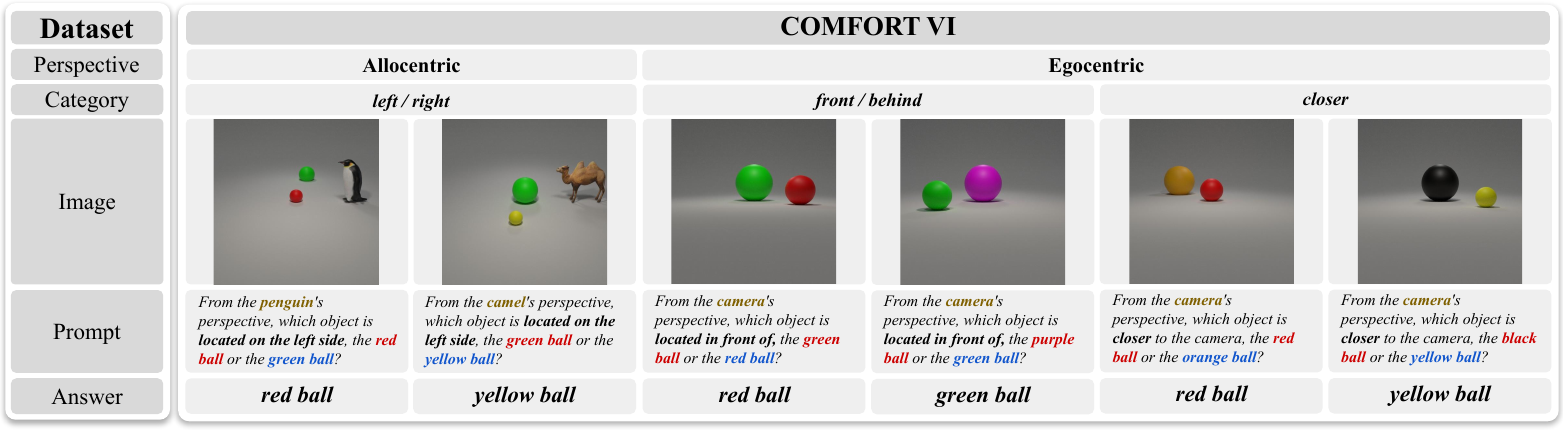}
    \caption{Examples of COMFORT VI. This dataset consists of three spatial relation categories: \textit{left/right} for allocentric spatial reasoning, and \textit{front/behind}, \textit{closer} for egocentric spatial reasoning.
    }
    \label{fig:dataset_vi}
\end{figure}

\noindent \textbf{COMFORT VI.} Figure~\ref{fig:dataset_vi} visualizes a spatial reasoning dataset designed for visual illusion scenarios, which was generated in the same Blender-based simulation environment as COMFORT\#.
The dataset consisted of the \textit{left/right} category for allocentric spatial reasoning and the \textit{front/behind} and \textit{closer} categories for egocentric spatial reasoning.  
Each scene contained two spheres of different colors and sizes placed in 3D space.  
For the \textit{left/right} category, a reference viewer was positioned to face the spheres from either the left or right side.  
To simulate visual illusion conditions, the sphere located farther from the camera was rendered significantly larger than the closer one, making the distant sphere appear larger in the image.

\begin{figure}[H]
    \centering
    \includegraphics[width=0.9\linewidth]{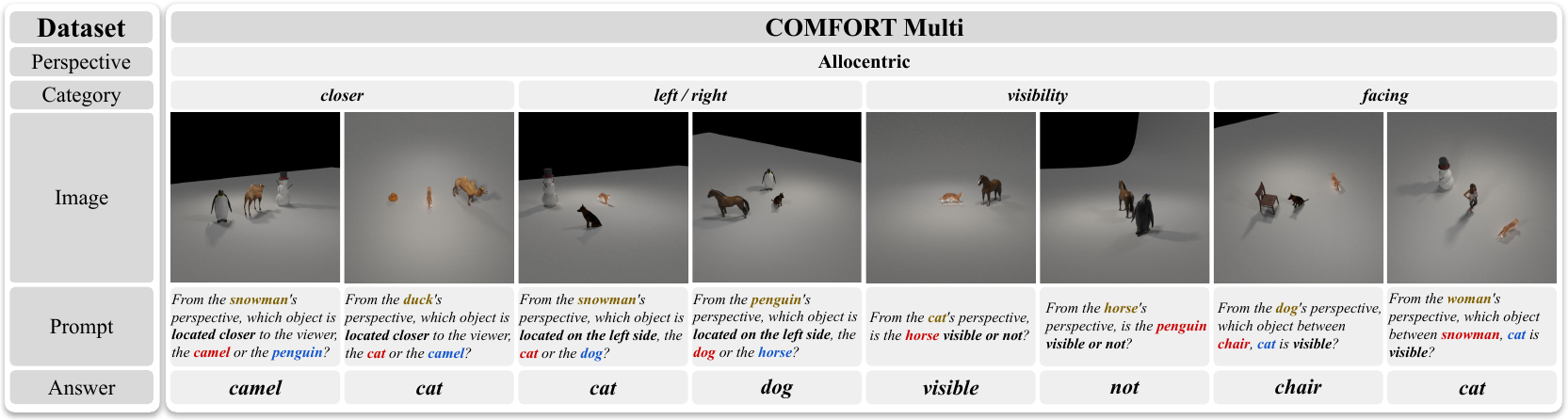}
    \caption{Examples of COMFORT Multi. This dataset consists of four spatial relation categories: \textit{closer}, \textit{left/right}, \textit{visibility} and \textit{facing}.}
    \vspace{-3mm}
    \label{fig:dataset_comfort_multi}
\end{figure}

\begin{figure}[H]
    \centering
    \includegraphics[width=0.75\linewidth]{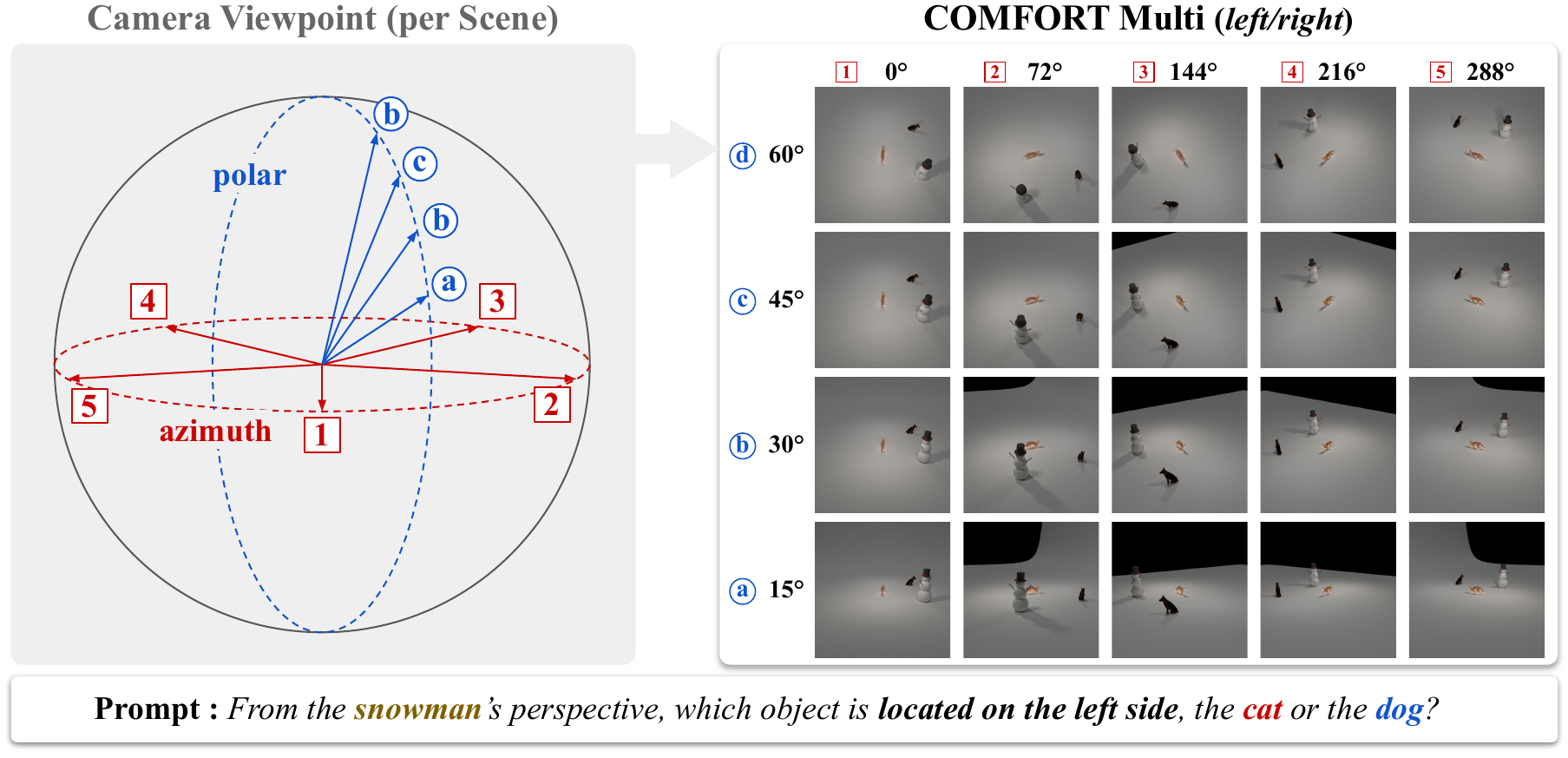}
    \caption{Dataset structure of COMFORT Multi. For each scene, on a spherical coordinate system centered at the scene, we captured 20 images by moving the camera viewpoint in steps of $15^\circ$ in polar angle and $72^\circ$ in azimuth.
    }
    \vspace{-3mm}
    \label{fig:dataset_comfort_multi_structure}
\end{figure}

\noindent \textbf{COMFORT Multi.} As illustrated in Figure~\ref{fig:dataset_comfort_multi}, COMFORT Multi was constructed in a Blender-based simulator to evaluate whether models can perform consistent allocentric spatial reasoning in the same scene from multiple viewpoints. 
For each category, we first created 10 scenes, and for each scene we extracted data from 20 viewpoints by varying the camera azimuth by $72^\circ$ and the polar angle by $15^\circ$ (see Figure~\ref{fig:dataset_comfort_multi_structure}).
All viewpoints captured from the same scene shared an identical prompt as input for inference.
The dataset was organized over the four categories: \textit{left/right}, \textit{closer}, \textit{visibility}, and \textit{facing}. 
The environment configuration defined for each category was identical to that of COMFORT\#.

\subsection{Perspective-Aware Spatial Reasoning Examples}
\begin{figure}[H]
    \centering
    \includegraphics[width=0.9\linewidth]{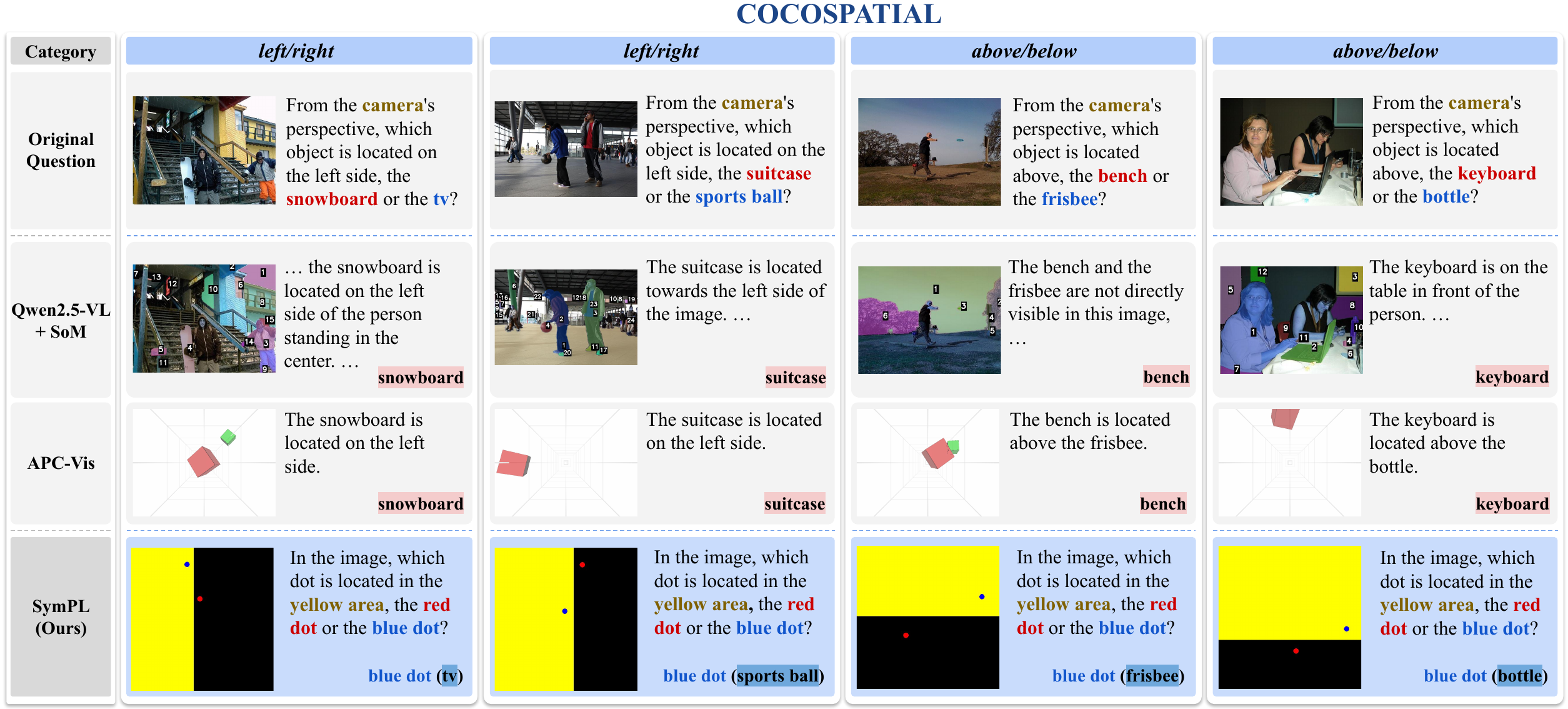}
    \caption{Egocentric spatial reasoning examples in the COCOSPATIAL dataset.
    }
    \vspace{-3mm}
    \label{fig:qualitative_result_coco}
\end{figure}

\begin{figure}[H]
    \centering
    \includegraphics[width=0.9\linewidth]{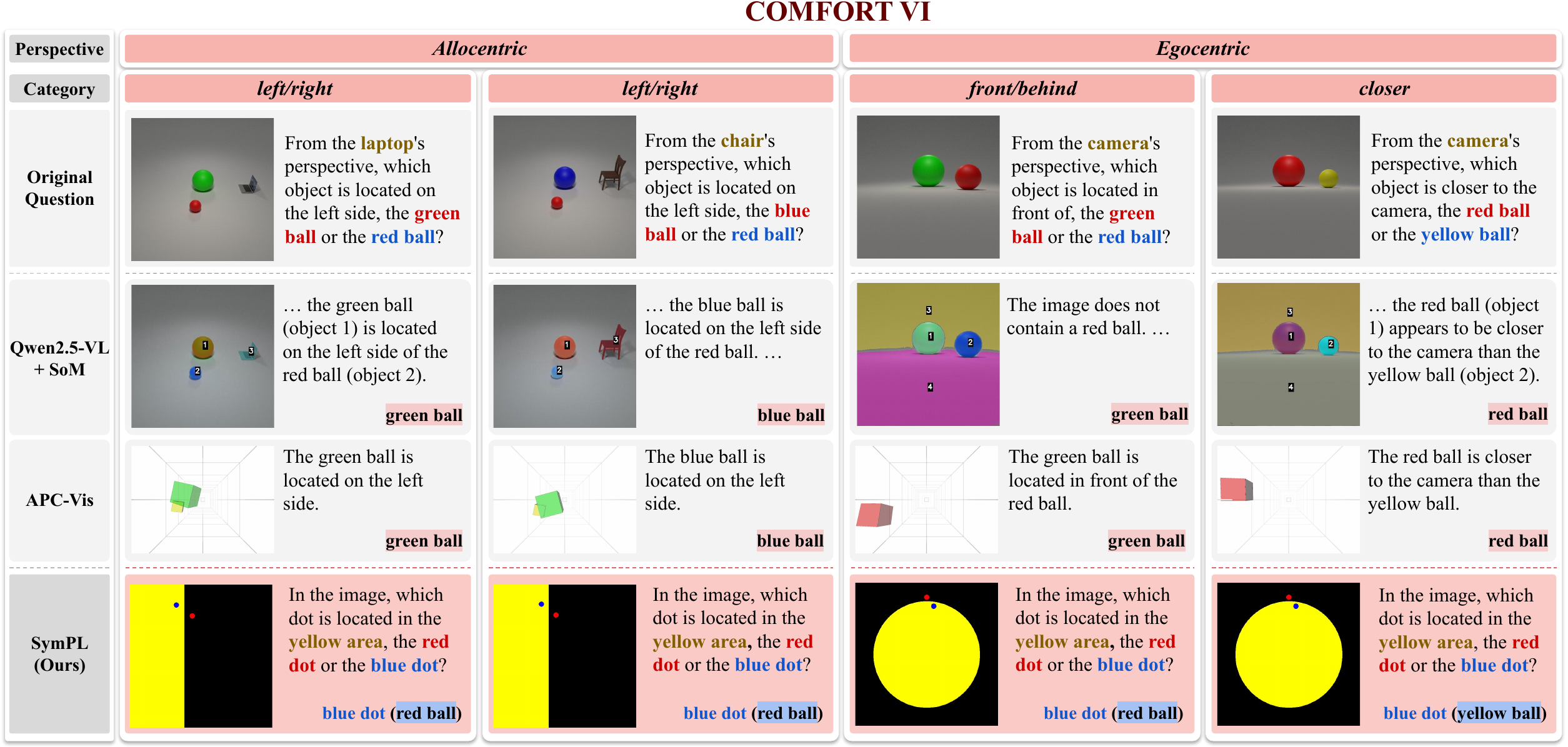}
    \caption{Perspective-aware spatial reasoning examples in the COMFORT VI dataset.
    }
    \label{fig:qualitative_result_vi}
\end{figure}

We evaluated the performance of the SymPL framework on a variety of spatial reasoning tasks using the COCOSPATIAL and COMFORT VI datasets, and the corresponding qualitative results were shown in Figure~\ref{fig:qualitative_result_coco} and~\ref{fig:qualitative_result_vi}.

\subsection{Ablation Study Details}

\subsubsection{Analysis of Each Key Factor}
To validate the effectiveness of the four key factors (projection, abstraction, bipartition, and localization), we conducted separate analyses for each of them.
All experiments were conducted separately for each of the five general purpose VLMs: Qwen2.5-VL~\cite{vlm_qwen2.5-vl}, GPT-5~\cite{vlm_gpt5}, LLaVA-NeXT~\cite{vlm_llavanext}, LLaVA-OneVision~\cite{vlm_llavaonevision}, and Molmo~\cite{vlm_molmo}.

\noindent \textbf{Projection.} 
To examine how viewpoint affected spatial reasoning performance, we considered two categories of spatial relations: \textit{left/right} and \textit{above/below}. 
For each scene, we collected images by moving the camera in $10^\circ$ increments: from a front view to a top view for the \textit{above/below} relations, and from a top view to a side view for the \textit{left/right} relations. 
For each category, we used data captured from 100 scenes.
In each experiment, we used the same question format: for \textit{left/right}, the question asked which object was located on the left side, and for \textit{above/below}, it asked which object was located above.

\begin{figure}[H]
  \centering

  \begin{subfigure}{0.48\linewidth}
    \centering
    \includegraphics[width=\linewidth]{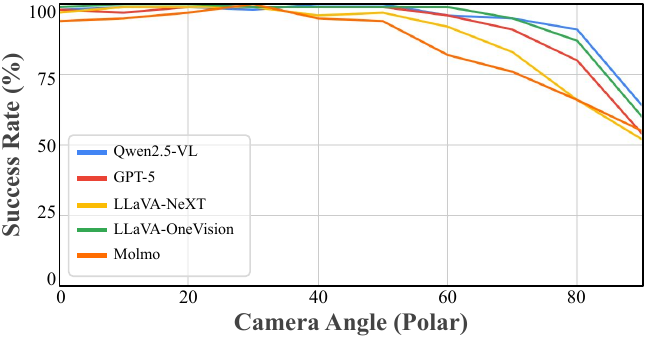}
    \captionsetup{margin={1.3em,0pt}}
    \caption{}
    \label{fig:ablation_projection:a}
  \end{subfigure}\hfill
  \begin{subfigure}{0.48\linewidth}
    \centering
    \includegraphics[width=\linewidth]{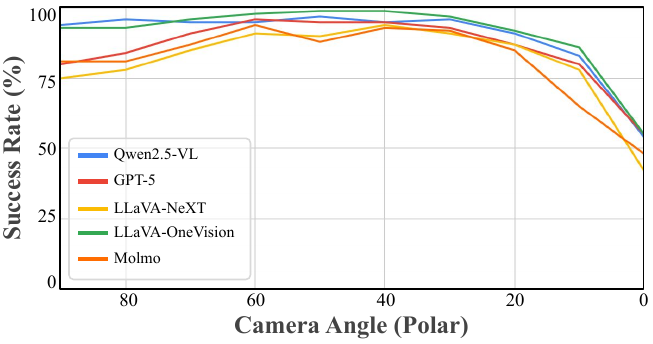}
    \captionsetup{margin={1.3em,0pt}}
    \caption{}
    \label{fig:ablation_projection:b}
  \end{subfigure}\hfill

  \caption{Spatial reasoning performance as a function of viewpoint position. The experiments were conducted on five general purpose VLMs. (a) shows the results for the \textit{above/below} category as the viewpoint moves from front to top view, and (b) shows the results for the \textit{left/right} category as it moves from top to side view.}
  \label{fig:ablation_projection}
\end{figure}

Figure~\ref{fig:ablation_projection} shows that, for both categories, performance improved when the viewpoint was approximately orthogonal to the plane in which the spatial relation was expressed.
Specifically, performance tended to increase as the viewpoint approached the front view for \textit{above/below} relations and the top view for \textit{left/right} relations. 
These findings indicated that the optimal viewpoint for spatial reasoning should depend on the type of spatial relation, and that this viewpoint was orthogonal to the plane in which the spatial relation was expressed.

\noindent \textbf{Abstraction.} 
We analyzed whether our abstraction factor had a meaningful effect on the reasoning performance of VLMs by evaluating performance under different forms of object representation.
The experiment was conducted on a simple scene where three objects were arranged in a row, using a \textit{closer} task that asked which object was closer to the rightmost object. 
For the analysis, we used 100 images for each of three conditions: the original image (original), the original image with segmentation masks annotated on the objects (seg mask), and an image in which the objects were abstracted into dot-shaped symbols (abstraction). 
When performing inferences with the abstraction image, we modified the prompt by replacing the original object names with the names of the abstracted symbols.

\begin{figure}[H]
    \centering
    \includegraphics[width=0.85\linewidth]{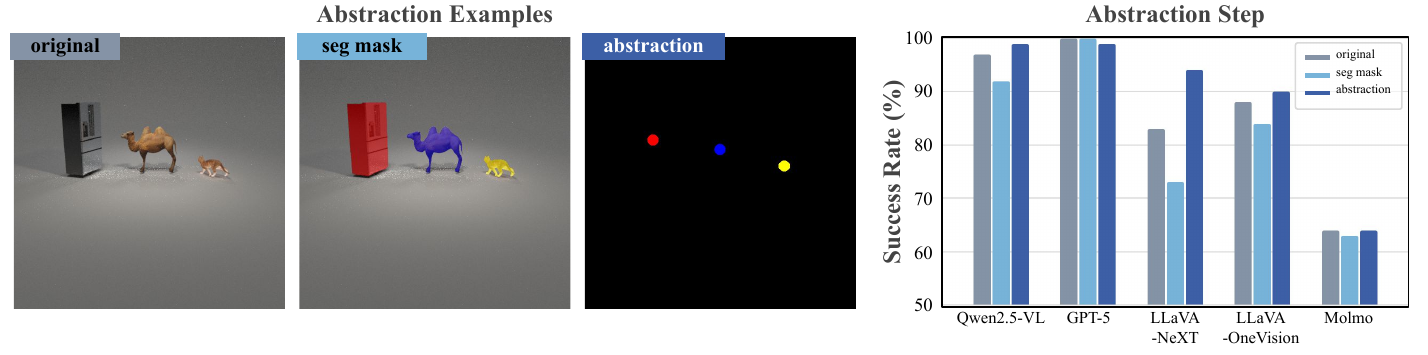}
    \caption{Spatial reasoning performance as a form of abstraction. The experiments were conducted on five general purpose VLMs. For each model, the results are shown from left to right in the following order: experiments on the original image (original), on the image with an annotated segmentation mask (seg mask), and on the abstracted image (abstraction).
    }
    \label{fig:ablation_abstraction}
\end{figure}

As shown in Figure~\ref{fig:ablation_abstraction}, we observed that, for most models, the reasoning performance on abstraction images tended to be higher. 
This suggested that abstract representations of objects had a more positive effect than the original images and the segmentation masks commonly used for visual prompting.

\noindent \textbf{Bipartition.}
We conducted an experiment to analyze whether partitioning the image was effective and how the number of partitions affected spatial reasoning performance. 
The experiment was performed using images generated by fixing the coordinates of each symbol and increasing the number of partitions from 1 to 4.
For each task, we used 100 images, and all tasks were formulated as questions asking which symbol was closer to the yellow dot.

\begin{figure}[H]
    \centering
    \includegraphics[width=\linewidth]{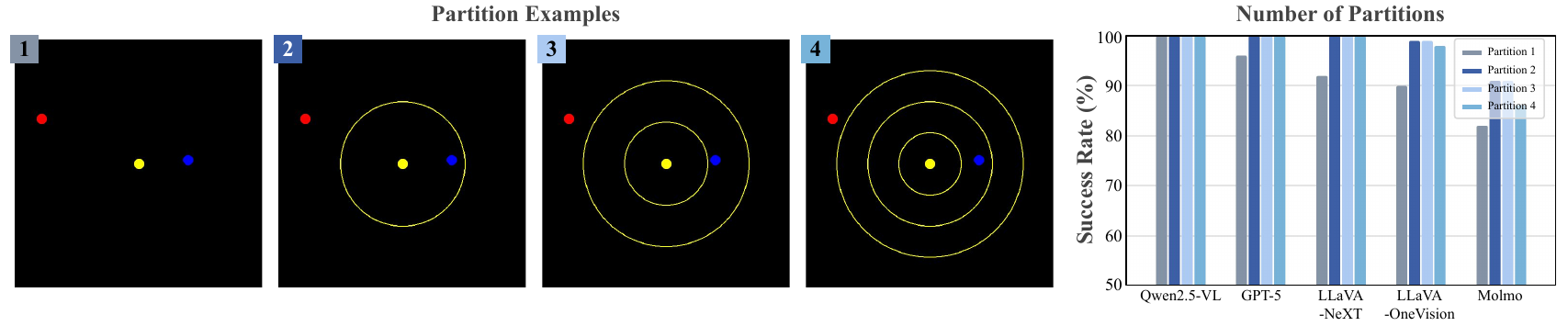}
    \caption{Spatial reasoning performance as a function of partition numbers. The experiments were conducted on five general purpose VLMs. For each model, the results are shown from left to right in the following order: images with 1, 2, 3, and 4 partitions.
    }
    \label{fig:ablation_bipartition}
\end{figure}

From the experimental results shown in Figure~\ref{fig:ablation_bipartition}, for most VLMs, reasoning performance was higher when the image was partitioned (partition 2, 3, or 4) than when it was not (partition 1). 
This suggested that dividing the space into partitions provided useful guidelines that help VLMs reason about spatial relations. 
Regarding the number of partitions, performance was highest when the image was divided into two or three partitions, while four partitions caused a slight performance drop in some models, although it still remained above the partition 1 case. 
Overall, these trends indicated that the presence of partitions had a more direct impact on spatial reasoning than the exact number of partitions.

\noindent \textbf{Localization.}
Following the ablation conducted in the \textit{bipartition} step, we further analyzed how the number of color-coded regions affected reasoning performance on the localization question by conducting additional experiments.
The experimental data consisted of images generated by fixing the coordinates of the symbols and varying the number of differently colored regions from 2 to 4. 
For each task, we used 100 images, and each task was formulated as a question asking for the color of the region in which one of the two symbols in the image was located.

\begin{figure}[H]
    \centering
    \includegraphics[width=0.85\linewidth]{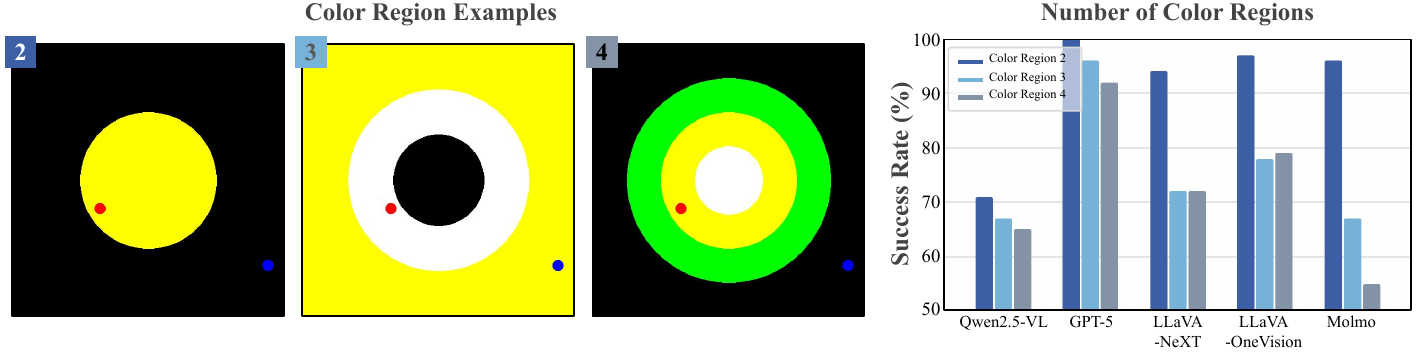}
    \caption{Spatial reasoning performance as a function of colored region numbers. The experiments were conducted on five general purpose VLMs. For each model, the results are shown from left to right in the following order: images with 2, 3, and 4 color regions.
    }
    \label{fig:ablation_localization}
\end{figure}

From the experimental results shown in Figure~\ref{fig:ablation_localization}, we observed that reasoning performance consistently decreased across all VLMs as the number of color-coded regions increased. 
This suggested that, when answering the localization problem, VLMs performed better when the image was divided into fewer color-coded regions. 
Taken together with the ablation results from the \textit{bipartition} step, these trends indicated that first splitting the space into two regions and then distinguishing them with different colors provided an effective way to solve the localization problem.

\subsubsection{Ablation on the Effectiveness of the Key Factor}
\begin{figure}[H]
  \centering

  \begin{subfigure}{0.24\linewidth}
    \centering
    \includegraphics[width=\linewidth]{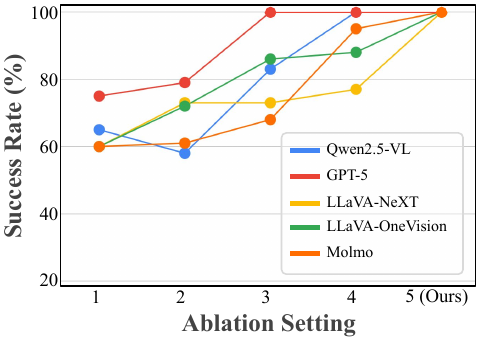}
    \captionsetup{margin={1.3em,0pt}}
    \caption{}
    \label{fig:ablation_result:a}
  \end{subfigure}\hfill
  \begin{subfigure}{0.24\linewidth}
    \centering
    \includegraphics[width=\linewidth]{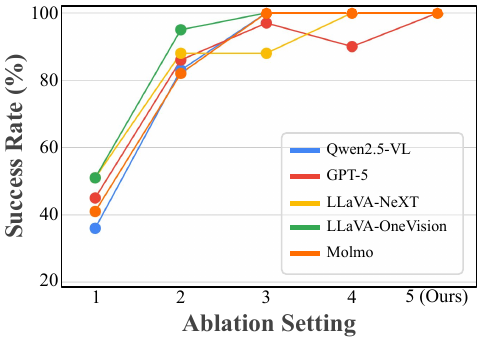}
    \captionsetup{margin={1.3em,0pt}}
    \caption{}
    \label{fig:ablation_result:b}
  \end{subfigure}\hfill
  \begin{subfigure}{0.24\linewidth}
    \centering
    \includegraphics[width=\linewidth]{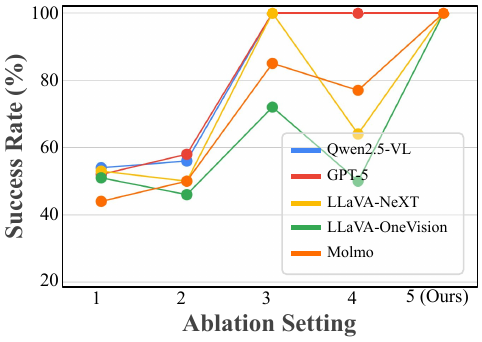}
    \captionsetup{margin={1.3em,0pt}}
    \caption{}
    \label{fig:ablation_result:c}
  \end{subfigure}\hfill
  \begin{subfigure}{0.24\linewidth}
    \centering
    \includegraphics[width=\linewidth]{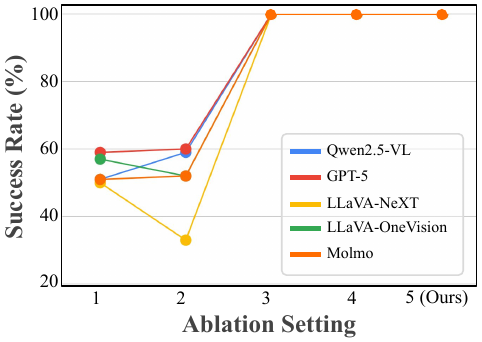}
    \captionsetup{margin={1.3em,0pt}}
    \caption{}
    \label{fig:ablation_result:d}
  \end{subfigure}

  \caption{Reasoning performance of five general purpose VLMs on the effectiveness of four key factors. (a) \textit{closer}, (b) \textit{left/right}, (c) \textit{visibility}, and (d) \textit{facing}. }
  \label{fig:ablation_result_detail}
\end{figure}

Figure~\ref{fig:ablation_result_detail} presents the results of the ablation study, evaluating the reasoning performance of VLMs on the effectiveness of four key factors.
In order, the graphs corresponded to the \textit{left/right}, \textit{closer}, \textit{visibility}, and \textit{facing} categories.  
We conducted experiments using five general purpose VLMs: Qwen2.5-VL, GPT-5, LLaVA-NeXT, LLaVA-OneVision, and Molmo.  
The experiments were conducted sequentially from Setting 1 to Setting 5. 
Each setting was obtained from the original allocentric question by sequentially applying the following factors: \textit{projection}, \textit{abstraction}, \textit{bipartition}, and \textit{localization}. 
Among these, the images in Setting 1 and Setting 2 were rendered directly in the simulation environment, while the remaining images were generated based on the object coordinates from the simulator. 

\begin{figure}[H]
    \centering
    \includegraphics[width=0.9\linewidth]{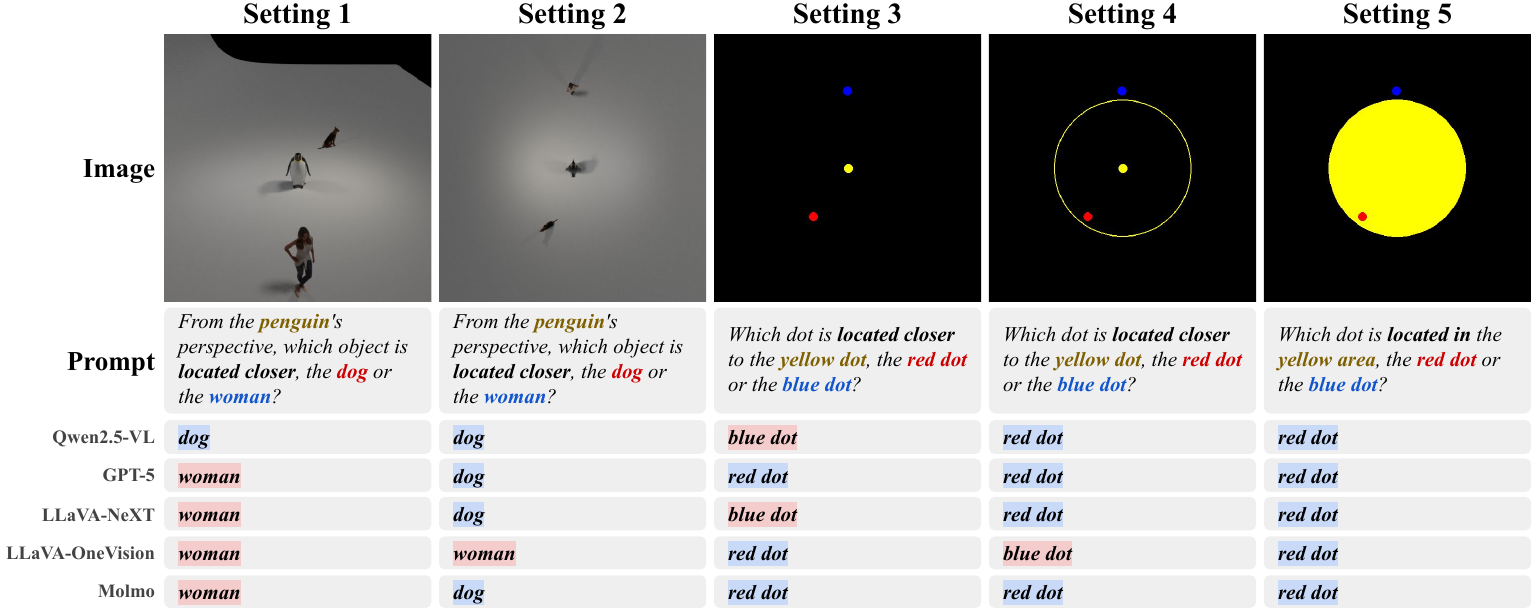}
    \caption{The qualitative results of the ablation study for the \textit{closer} category.}
    \label{fig:supplimentary_ablation_closer}
\end{figure}

\begin{figure}[H]
    \centering
    \includegraphics[width=0.9\linewidth]{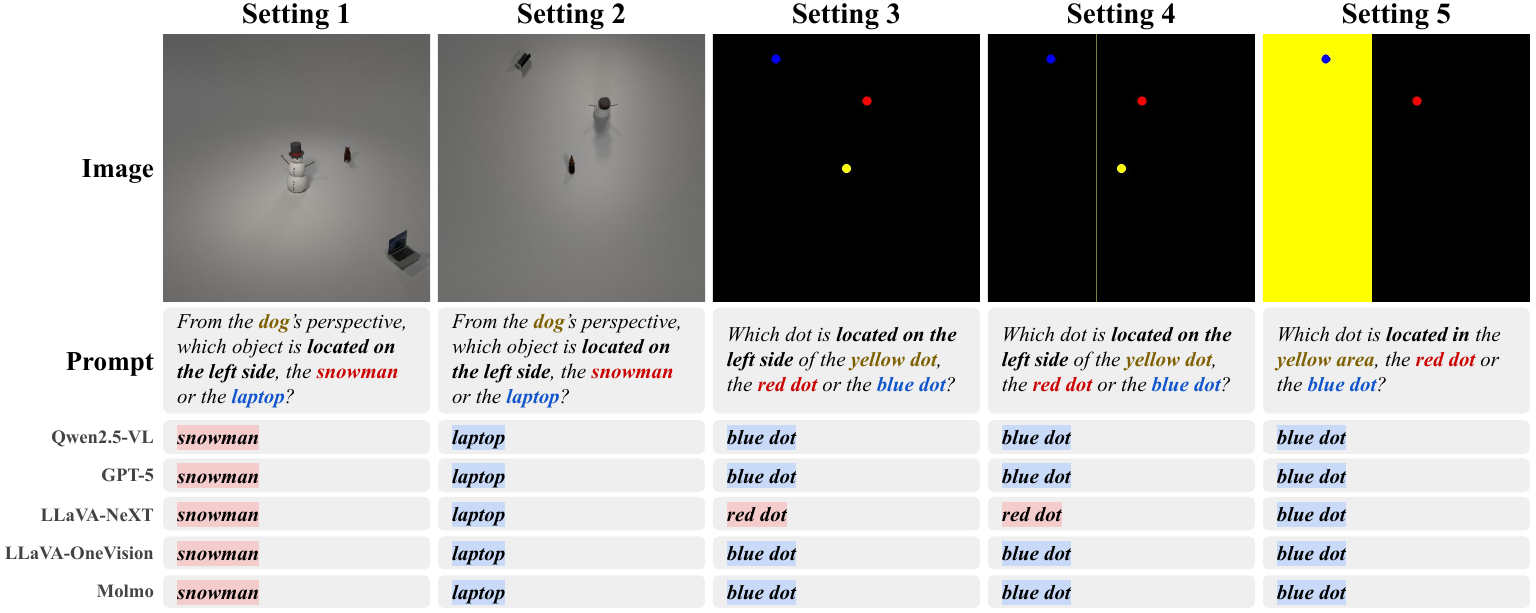}
    \caption{The qualitative results of the ablation study for the \textit{left/right} category.}
    \label{fig:supplimentary_ablation_left_right}
\end{figure}

\begin{figure}[H]
    \centering
    \includegraphics[width=0.9\linewidth]{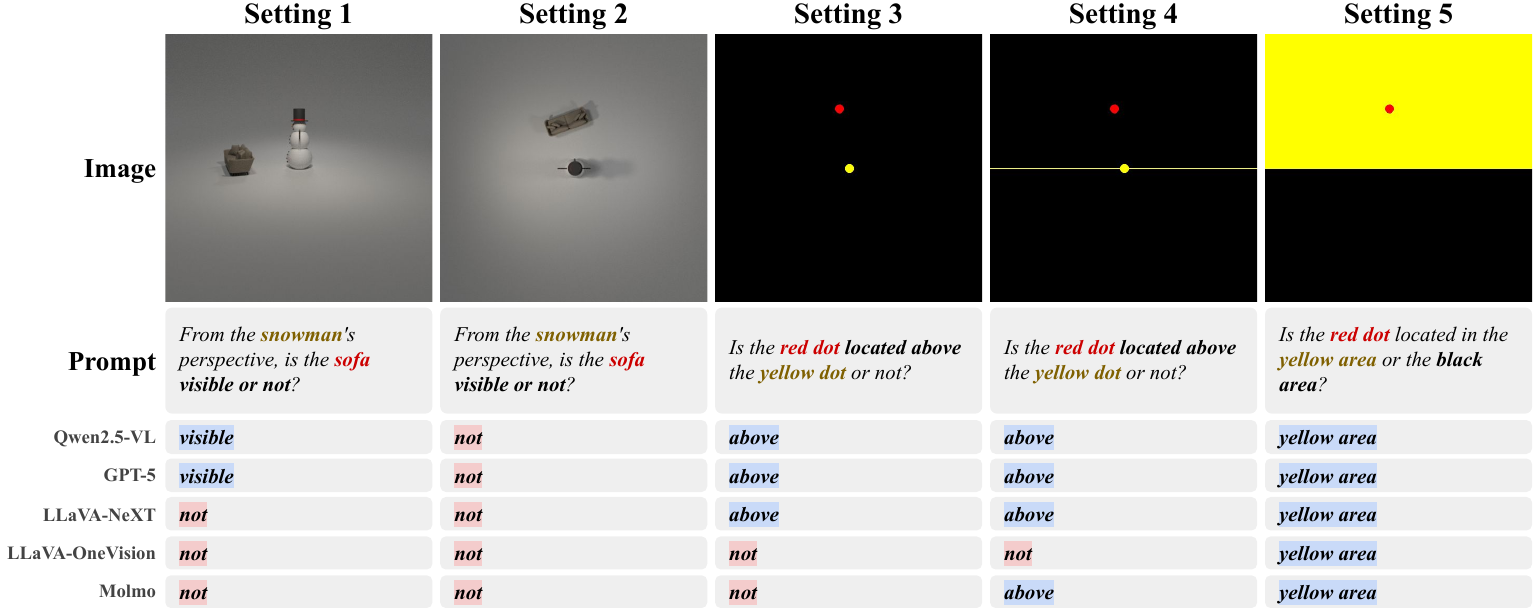}
    \caption{The qualitative results of the ablation study for the \textit{visibility} category.}
    \label{fig:supplimentary_ablation_visibility}
\end{figure}

\begin{figure}[H]
    \centering
    \includegraphics[width=0.9\linewidth]{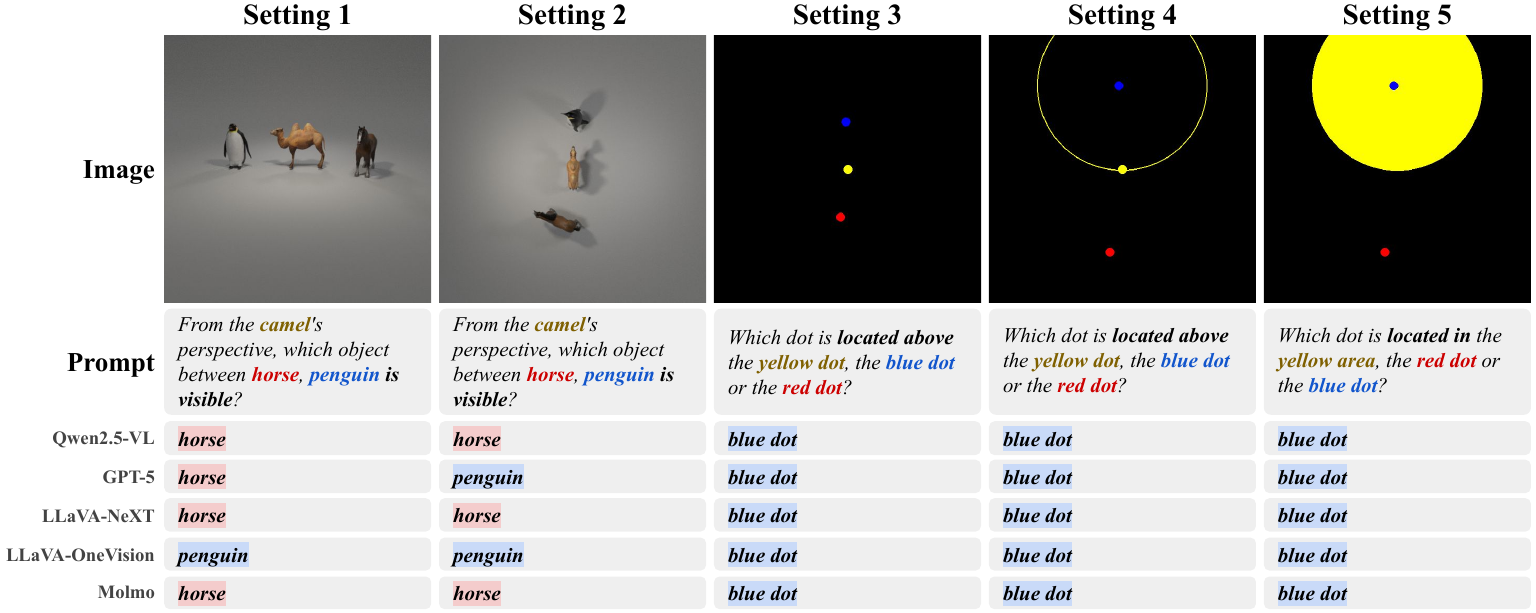}
    \caption{The qualitative results of the ablation study for the \textit{facing} category.}
    \label{fig:supplimentary_ablation_facing}
\end{figure}

From the experimental results, we observed that the reasoning performance, which was low in Setting 1 (allocentric question), gradually increased as the key factors were incrementally introduced, and eventually reached 100\% in Setting 5 (symbolic-layout question). 
This trend consistently appeared regardless of the type of general purpose VLM. 
These results showed that the robust reasoning capability of the proposed symbolic-layout question, based on the four key factors, was generally preserved across diverse VLMs.
Examples and qualitative results for each setting are shown in Figure~\ref{fig:supplimentary_ablation_closer}, ~\ref{fig:supplimentary_ablation_left_right}, ~\ref{fig:supplimentary_ablation_visibility}, and~\ref{fig:supplimentary_ablation_facing}.

\subsection{Applying SymPL to Other VLM}
We evaluated whether applying the SymPL framework with a VLM other than Qwen2.5-VL also led to consistent performance gains. 
To do this, we replaced Qwen2.5-VL with GPT-5 throughout the reasoning pipeline and analyzed allocentric spatial reasoning performance.

\begin{figure}[H]
  \centering

  \begin{subfigure}{0.4\linewidth}
    \centering
    \includegraphics[width=\linewidth]{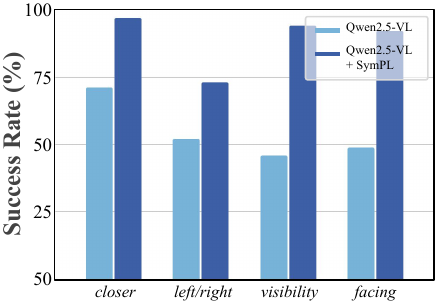}
    \captionsetup{margin={1.3em,0pt}}
    \caption{}
    \label{fig:sympl_application:a}
  \end{subfigure}
  \hspace{0.05\linewidth}
  \begin{subfigure}{0.4\linewidth}
    \centering
    \includegraphics[width=\linewidth]{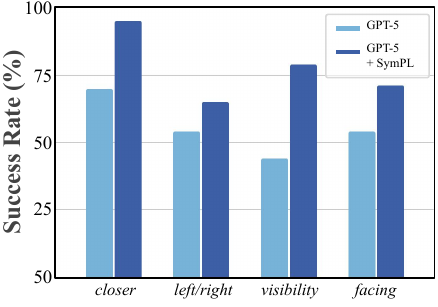}
    \captionsetup{margin={1.3em,0pt}}
    \caption{}
    \label{fig:sympl_application:b}
  \end{subfigure}\hfill

  \caption{Allocentric spatial reasoning performance with applying SymPL to GPT-5. (a) shows the performance of Qwen2.5-VL before and after applying SymPL, and (b) shows the corresponding results for GPT-5.}
  \label{fig:sympl_application}
\end{figure}

According to the results in Figure~\ref{fig:sympl_application}, similar to the trend observed with Qwen2.5-VL, applying the SymPL framework with GPT-5 improved performance across all categories. 
These results indicate that the proposed framework is not limited to the Qwen2.5-VL model alone.

\clearpage

\section{Reasoning Prompt}
The SymPL framework includes a stepwise procedure that utilizes the reasoning capability of the VLM at each step.  
This procedure consists of five steps: target object selection, detection refinement, reference viewer selection, category determination, and symbolic-layout reasoning.  
The text prompts used for reasoning in each step are as follows.

\begin{tcolorbox}[
coltitle=black, 
fonttitle=\footnotesize\ttfamily, 
fontupper=\footnotesize\ttfamily,
colback=gray!5, 
colframe=gray!40, 
title=\textbf{Prompt - Target Object Selection}]

\textbf{\# Situation Description}

Given a spatial reasoning question, please return all the words the represent the entities that are included in the question.

\bigskip

\textbf{\# Example} 

\textbf{[Question]} From the old man's perspective, is the person wearing a hat on the left of the green car?

\textbf{[Detect]} [old man, person wearing a hat, green car]

\bigskip

\textbf{\# Your Task}

Now, given the question below, please identify the entities that are included in the question.

All the results return as a format [Detect] [object\_1, object\_2, ...].

\bigskip

\textbf{[Question]} \{question\}

\textbf{[Detect]}
\end{tcolorbox}

\begin{tcolorbox}[
coltitle=black, 
fonttitle=\footnotesize\ttfamily, 
fontupper=\footnotesize\ttfamily,
colback=gray!5, 
colframe=gray!40, 
title=\textbf{Prompt - Detection Refinement}]

The input images are the cropped regions from the original image that correspond to description : `{category}'.

Look at each of these images and select the one that best matches description : `{category}'.

\bigskip

Your response should return only the index number of the image you selected.

Note : If multiple images are considered a match, select the one with the lowest index number.
\end{tcolorbox}

\begin{tcolorbox}[
coltitle=black, 
fonttitle=\footnotesize\ttfamily, 
fontupper=\footnotesize\ttfamily,
colback=gray!5, 
colframe=gray!40, 
title=\textbf{Prompt - Reference Viewer Selection}]

\textbf{\# Situation Description}

Given a question about spatial reasoning, we want to extract the **perspective** of the question.

If the question is from the camera's perspective or cannot mention the perspective, return ++camera++.
Never return anything else.

\bigskip

\textbf{\# Example} 

\textbf{[Question]} If I stand at the shepherd's position facing where it is facing, is the sheep visible or not

\textbf{[Perspective]} ++shepherd++

\bigskip

\textbf{\# Your Task}

Given the question below, please specify the **perspective** from which the question is asked.

After ``[Perspective]'' at the end of this prompt, you must return the answer for the base object in the ``object\_name'' field, following the format : ++object\_name++

``object\_name'' must be selected only from the [Option] list provided below.

Never return any answer outside of these options.

Just include ++ in front of and behind of the selected ``object\_name'' candidate. Never change anything else.

\bigskip

\textbf{[Question]} \{question\}

\textbf{[Options]} \{obj\_str\}, camera

\textbf{[Perspective]}
\end{tcolorbox}

\begin{tcolorbox}[
coltitle=black, 
fonttitle=\footnotesize\ttfamily, 
fontupper=\footnotesize\ttfamily,
colback=gray!5, 
colframe=gray!40, 
title=\textbf{Prompt - Category Determination}]

\textbf{\# Situation Description}

Given a question about spatial reasoning, we want to extract the category of the question.

The words inside ** ** in the [Question] are the key elements of that [Category].

Depending on the expression, words such as ``visible'' or ``facing'' may appear in [Question]. 

However, the mere presence of these words does not determine that [Category] should be ``visibility'' or ``facing.''

Refer to the parts highlighted with ** ** in the examples and select the most appropriate [Category].

\bigskip

\textbf{\# Example} 

\textbf{[Question]} If I stand at the man in cowboy hat's position facing where it is facing, is the bus stop **on the left or right** of me?

\textbf{[Category]} --left\_right--

\bigskip

\textbf{\# Your Task}

Given the question below, please specify the category from which the question is asked.

You must return in the format: [Category] --category\_name--

``object\_name'' is selected from [Options] below. 

Never return a response that is not included in the given options.

Never change the format and capitalization from the option when returns response.

\bigskip

\textbf{[Question]} \{question\}

\textbf{[Options]} visibility / left\_right / facing / closer / above\_below / front\_behind

\textbf{[Category]}
\end{tcolorbox}

\begin{tcolorbox}[
coltitle=black, 
fonttitle=\footnotesize\ttfamily, 
fontupper=\footnotesize\ttfamily,
colback=gray!5, 
colframe=gray!40, 
title=\textbf{Prompt - Symbolic-Layout Reasoning (left/right1, visibility)}]

This is an image of a simple 2D Scene.

\bigskip

\textbf{\# Task}

Based on the image, please answer the following question.

\bigskip

\textbf{[Question]} In the image, is the \{obj\} dot located in the `yellow' area or the `black' area?

\bigskip

Please only return the answer.

\end{tcolorbox}

\begin{tcolorbox}[
coltitle=black, 
fonttitle=\footnotesize\ttfamily, 
fontupper=\footnotesize\ttfamily,
colback=gray!5, 
colframe=gray!40, 
title=\textbf{Prompt - Symbolic-Layout Reasoning (left/right2, closer, facing, front/behind, above/below)}]

This is an image of a simple 2D Scene.

\bigskip

\textbf{\# Task}

Based on the image, please answer the following question.

\bigskip

\textbf{[Question]} In the image, which dot is located in the `yellow' area, the \{obj\_1\} dot or the \{obj\_2\} dot?

\bigskip

Please only return the answer.

\end{tcolorbox}

\clearpage
{
    \small
    \bibliographystyle{ieeenat_fullname}
    \bibliography{main}
}
